\documentclass[pdftex]{article}

    \usepackage[dandb, final]{neurips_2025}

\usepackage[utf8]{inputenc} 
\usepackage[T1]{fontenc}    
\usepackage{hyperref}       
\usepackage{url}            
\usepackage{subcaption} 
\usepackage{booktabs}       
\usepackage{amsfonts}       
\usepackage{amsmath}
\usepackage{nicefrac}       
\usepackage{microtype}      
\usepackage{xcolor}         
\usepackage{rotating}
\usepackage{graphicx}
\usepackage{multirow}
\usepackage{makecell}
\usepackage[ruled,vlined]{algorithm2e}
\usepackage{pythonhighlight}
\usepackage{tcolorbox}
\usepackage{hyperref}
\definecolor{lightblue}{RGB}{0,174,239} 
\newcommand{\good}[1]{\textcolor{green!70!black}{#1}}
\newcommand{\bad}[1]{\textcolor{red!70!black}{#1}}
\newcommand{\neutral}[1]{\textcolor{gray!70!black}{#1}}

\definecolor{lightblue}{RGB}{0,150,255} 

\title{Alchemist: Turning Public Text-to-Image Data \\ into Generative Gold}

\author{%
  Valerii Startsev\thanks{Equal contribution.} \\
  Yandex Research, HSE \\
  \And
  Alexander Ustyuzhanin$^*$\\
  Yandex \\
  \AND
  Alexey Kirillov \\
  Yandex, MSU \\
  \And
  Dmitry Baranchuk \\
  Yandex Research \\
  \And
  Sergey Kastryulin \\
  Yandex Research \\
}

\begin{document}
\maketitle
\vspace{-25pt}

\begin{center}
\href{https://huggingface.co/collections/yandex/alchemist-6825f7a16cbcc71128ee525f}{\color{lightblue}\url{https://huggingface.co/datasets/yandex/alchemist}}
\end{center}
\vspace{-5pt}

\begin{figure}[h!]
    \centering
    \includegraphics[width=0.99\textwidth]{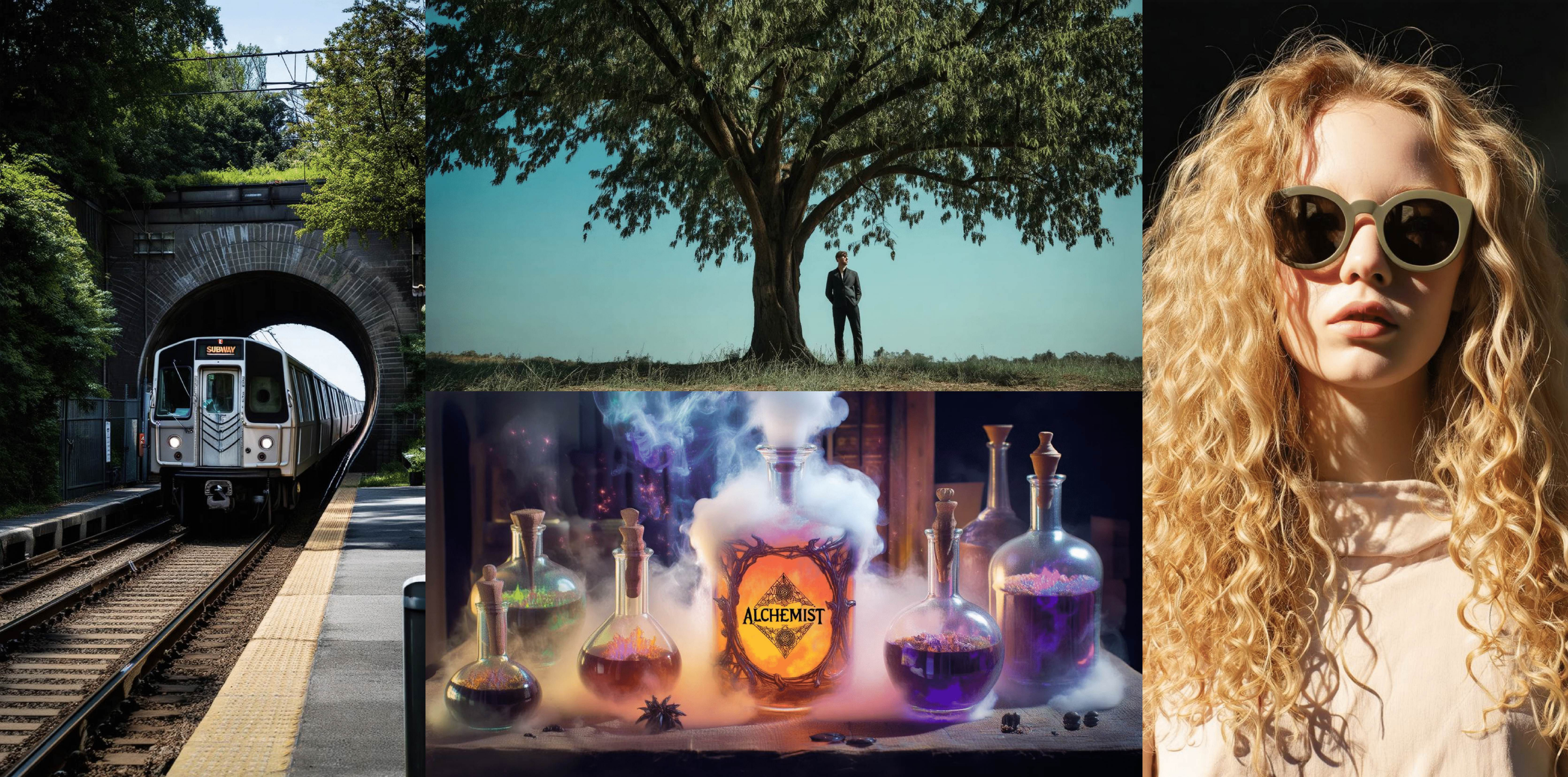} 
    \caption{{\small Images generated by Stable Diffusion 3.5 Large fine-tuned on Alchemist, demonstrating enhanced aesthetic quality and complexity while maintaining prompt adherence.}}
    \label{fig:mosaic}
\end{figure}

\vspace{-5pt}
\begin{abstract}{
\vspace{-5pt}

Pre-training equips text-to-image (T2I) models with broad world knowledge, but this alone is often insufficient to achieve high aesthetic quality and alignment. 
Consequently, supervised fine-tuning (SFT) is crucial for further refinement. 
However, its effectiveness highly depends on the quality of the fine-tuning dataset.
Existing public SFT datasets frequently target narrow domains (e.g., anime or specific art styles), and the creation of high-quality, general-purpose SFT datasets remains a significant challenge.
Current curation methods are often costly and struggle to identify truly impactful samples.
This challenge is further complicated by the scarcity of public general-purpose datasets, as leading models often rely on large, proprietary, and poorly documented internal data, hindering broader research progress.
This paper introduces a novel methodology for creating general-purpose SFT datasets by leveraging a pre-trained generative model as an estimator of high-impact training samples. 
We apply this methodology to construct and release \textbf{Alchemist}, a compact (3,350 samples) yet highly effective SFT dataset. 
Experiments demonstrate that Alchemist substantially improves the generative quality of five public T2I models while preserving diversity and style. 
Additionally, we release the fine-tuned models' weights to the public}.
\end{abstract}

\section{Introduction}
\label{sec:introduction}

Generative text-to-image (T2I) models, such as DALL-E 3 \cite{betker2023improving}, Imagen 3 \cite{baldridge2024imagen3}, and Stable Diffusion 3 \cite{esser2024scaling}, have demonstrated remarkable advancements in synthesizing high-fidelity and diverse images from textual descriptions. 
These models, typically pre-trained on vast internet-scale datasets, are applied in creative industries, for content generation, and in scientific visualizations.
Despite their capabilities, the continuous pursuit of enhanced generative quality and better alignment with user intent remains a central research focus.

Supervised Fine-Tuning (SFT) has become a vital technique for adapting these pre-trained foundation models, whether to specialize them for particular domains or aesthetics, or to broadly elevate their general generative performance.
However, the success of SFT is critically dependent on the quality and composition of the fine-tuning dataset.
Current practices for SFT dataset curation often rely on extensive manual human selection.
This process is not only costly and challenging to scale but can also be surprisingly ineffective.
The specific characteristics of text-image pairs that render a sample "good" for SFT -- that is, likely to maximally boost general model quality -- are frequently subtle, not obvious, and difficult for humans to consistently verbalize or identify. 
Alternative approaches, such as filtering large web datasets with simple heuristics or employing synthetic data generation, have their own limitations in efficiently targeting high-impact samples or ensuring quality and diversity without introducing new biases.

These methodological challenges are compounded by a significant scarcity of publicly available, general-purpose SFT datasets explicitly designed to broadly enhance T2I models.
While numerous domain-specific fine-tuning datasets exist, they serve niche applications rather than general quality improvement.
Furthermore, several recent state-of-the-art models (e.g., Emu \cite{dai2023emu}, PixArt-$\alpha$ \cite{chen2023pixart}, Kolors \cite{team2024kolors}, SANA \cite{xie2024sana}, YaART \cite{kastryulin2024yaart}) report using internal datasets for their SFT stages. 
These datasets, however, remain closed-source and are often described with insufficient detail in publications, severely limiting the research community's ability to replicate findings, understand their construction principles, or develop comparable open resources. 
This lack of accessible, well-characterized, general-purpose SFT datasets impedes broader progress in systematically improving T2I models.

To address these challenges, we propose a novel approach that leverages the intrinsic understanding of a pre-trained generative model to more effectively guide the SFT dataset creation process.
Our core idea is that a pre-trained generative model can itself serve as an  estimator of data quality, pinpointing samples most likely to contribute positively to the fine-tuning objective and maximize generative improvements in downstream models.
To demonstrate the practical utility of this methodology, we created the Alchemist dataset and subsequently used it to fine-tune five publicly available text-to-image models, the improved weights of which we release as part of our contributions.

This work aims to provide the first open general-purpose alternative to proprietary fine-tuning pipelines, enabling reproducible research and commercial applications.

We present the following contributions:
\begin{itemize}
    \item A principled methodology for curating high-quality, general-purpose SFT datasets by leveraging a pre-trained generative model to identify samples that maximize post-SFT model improvement.
    \item Alchemist, a compact (3,350 samples) yet highly effective SFT dataset constructed via our methodology, significantly enhances text-to-image generation quality while maintaining output diversity and style.
    \item Open-sourced, fine-tuned weights for five publicly available text-to-image models, demonstrating performance gains over their baselines after SFT with Alchemist.
\end{itemize}

The remainder of this paper details our methodology for dataset creation, presents experimental results showcasing its effectiveness, and discusses the outcomes and limitations of our research.


\section{Related Work}
\label{sec:related_work}

\paragraph{Supervised Fine-Tuning of Text-to-Image Models.}

Early text-to-image models like DALL-E \cite{ramesh2021zero} and Latent Diffusion Models (LDMs) \cite{rombach2022high} were primarily pre-trained on vast, uncurated web-scale datasets (e.g., LAION-5B \cite{schuhmann2022laion}), focusing on general generative capabilities without specific SFT stages. 
A significant advancement came with Emu \cite{dai2023emu}, which demonstrated that an SFT stage on a smaller, high-quality, curated dataset substantially improved instruction following and aesthetic quality. 
Subsequently, SFT or similar refinement stages became standard in state-of-the-art models. 
For instance, PixArt-$\alpha$ \cite{chen2023pixart} enhances outputs through training on data with higher aesthetic quality, boosting the training efficiency. 
Later works \cite{team2024kolors, xie2024sana, kastryulin2024yaart} also employ multi-stage training including fine-tuning on aesthetically filtered data. 
This trend highlights crucial role of SFT in achieving high-quality and controllable image generation.

\paragraph{Supervised Fine-Tuning Datasets for Text-to-Image Models.}

Publicly available, general-purpose SFT datasets for text-to-image models remain limited. 
LAION-Aesthetics \cite{laion-aesthetics}, derived from LAION-5B \cite{schuhmann2022laion} by filtering for predicted aesthetic scores, is widely used.
However, its quality is often considered inferior compared to closed source datasets.
While more recent efforts, such as LAION-Aesthetics V2 \cite{laion-aesthetics-v2}, aim to improve upon this, a meticulously verified, general-purpose public SFT dataset is largely absent.
In contrast, domain-specific SFT datasets are more common, such as the Danbooru dataset \cite{danbooru2021} for anime-style generation and WikiArt dataset \cite{WikiArt_OnlineCollection} for classical and modern art generation. 
These datasets achieve strong performance within their specific domains but typically at the cost of the model's broader generative abilities, causing it to overfit to the narrow domain of the SFT data.
The scarcity of high-quality, general-purpose public SFT datasets motivates our work.

\paragraph{Quality Assessment of Text-to-Image Models.}

Evaluating text-to-image generation quality is complex. 
Automated metrics like Fr\'echet Inception Distance (FID) \cite{heusel2017gans} and Inception Score (IS) \cite{salimans2016improved}, while common, often correlate poorly with human perception \cite{borji2019_pros_and_cons_of_gan_metrics, sajjadi2018_precision_recall_metric}.
Consequently, more comprehensive evaluations rely on human assessment. 
Studies for models like Imagen \cite{saharia2022photorealistic} and Parti \cite{yu2022scaling}, for example, involved human raters evaluating photorealism, text-image alignment, absence of artifacts, and compositionality.
Standardized prompt sets such as DrawBench \cite{saharia2022photorealistic} and T2I-CompBench \cite{huang2023t2i} facilitate structured comparison. 
In this work we provide some automated metrics while building main conclusions based on carefully designed multi-aspect human-based evaluations.

\section{Dataset Formation}
\label{sec:method}


Our goal is to create a general-purpose supervised fine-tuning  dataset capable of significantly enhancing the generative quality of pre-trained text-to-image (T2I) models while preserving their  diversity in content, composition, and style. 
To achieve this, we introduce a multi-stage filtering pipeline designed to create a small set of exceptionally high-quality samples from a vast pool of uncurated internet data. 
A core principle of our methodology involves leveraging a pre-trained diffusion model as a sophisticated estimator in the final filtering stage to identify text-image pairs with the highest potential to boost downstream SFT performance. 
This section details our pipeline, the effectiveness of which is demonstrated in Section \ref{sec:experiments}.

\paragraph{Overview.}

\begin{figure}[tb]
    \centering
    \includegraphics[width=0.99\textwidth]{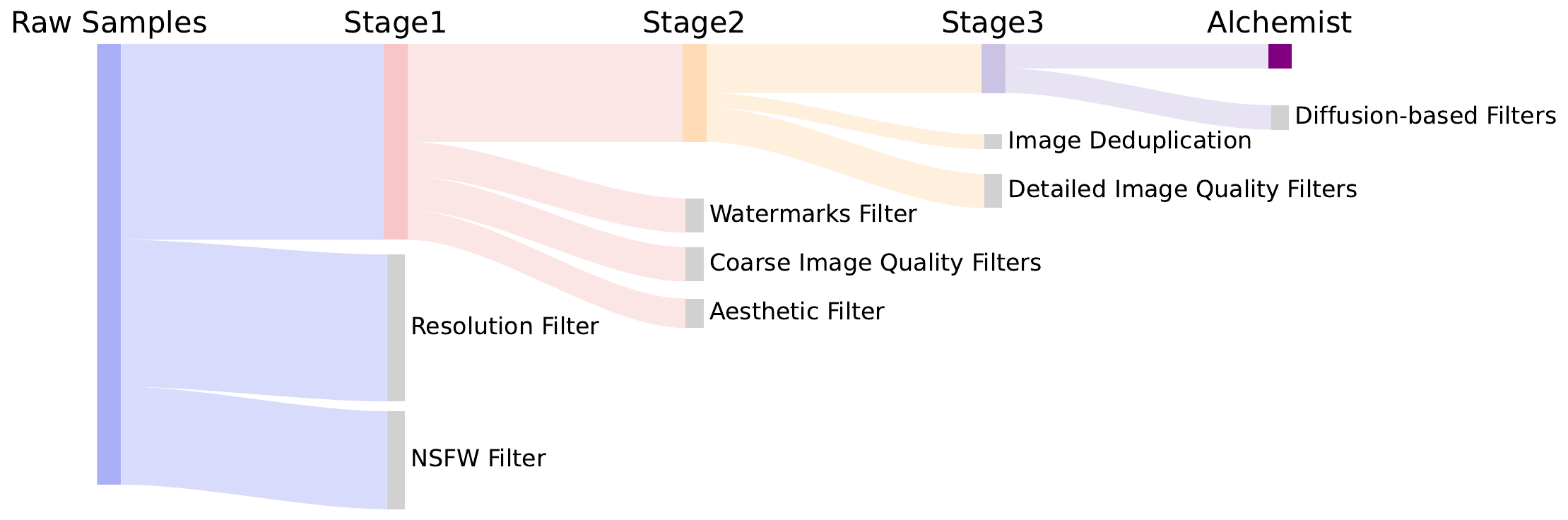} 
    \caption{{\small Overview of the multi-stage image filtering pipeline. Beginning with a web-scale collection of raw data, the pipeline sequentially filters images to isolate a high-quality subset optimally suited for supervised fine-tuning of text-to-image models.}}
    \label{fig:data_filter_pipeline}
    \vspace{-5px}
\end{figure}

The dataset construction process starts from a vast, diverse pool of $\mathcal{O}$(10 billion) images aggregated from  web-scraped sources.
Some dataset curation pipelines for image-text models discussed in literature \cite{gadre2023datacomp} impose text-based filtering at the initial stages, discarding samples with poorly structured, noisy or semantically misaligned captions. 
While this approach mitigates low-quality training pairs, we argue that it is increasingly restrictive given recent advances in multi-modal captioning models.
Early text filtering eliminates potentially valuable visual content that could be re-captioned with synthetic texts.
Instead, we compose our data curation pipeline as purely image-based. 
The relatively high-quality set of data that passes first filtering stages is further filtered using diffusion-based sample quality estimator and then captioned with a Vision-Language Model (VLM) to obtain the final SFT dataset.
Figure \ref{fig:data_filter_pipeline} provides an overview of the dataset formation pipeline.

\paragraph{Stage 1: Foundational Safety and Resolution Filtering.}
The first filtering stage addresses basic image requirements. 
We discarded images identified as containing Not-Safe-For-Work (NSFW) content through an automated classifier. 
Subsequently, we applied a resolution filter, retaining only those images with an area exceeding $1024 \times 1024$ px. 
This step ensures that the candidate pool consists of sufficiently high-resolution and safe visual content for subsequent processing.

\paragraph{Stage 2: Coarse-Grained Quality Assessment and Filtering.}
Following the foundational filters, we employed a suite of lightweight binary classifiers for rapid, coarse-grained quality assessment. 
These classifiers were trained to identify and remove images exhibiting severe degradation, watermarks, noticeable compression artifacts, significant motion blur, or low aesthetic appeal. 
The training data for these classifiers was derived from public Image Quality Assessment (IQA) \cite{koniq10k, jinjin2020pipal} and Image Aesthetics Assessment (IAA) \cite{he2022_TAD, murray2012ava} datasets. 
Classification thresholds were manually calibrated to aggressively remove the worst-quality examples. 
The first two stages significantly reduced the dataset size, yielding approximately one billion images for further processing.

\vspace{-5px}
\paragraph{Stage 3: Deduplication and Fine-Grained Quality Refinement.}

With a more manageable dataset size, we applied more computationally intensive methods. 
First, to enhance visual diversity, we performed image deduplication by computing SIFT-like local features \cite{lowe2004distinctive}, clustering images by similarity, and retaining only one representative (highest preliminary quality score) per cluster. 
Second, for fine-grained perceptual quality assessment, we utilized the TOPIQ no-reference IQA model \cite{chen2024topiq}.
After empirical studies to balance quality with domain diversity (avoiding bias towards narrow domains like architecture and interiors), we set the TOPIQ threshold to $> 0.71$. 
This step isolated images with minimal distortions and artifacts while preserving broad thematic coverage, resulting in approximately 300 million high-quality images.

\vspace{-5px}
\paragraph{Perceptual and Compositional Refinement.}

\begin{figure} 
\centering
\begin{minipage}{0.95\textwidth} 
\begin{algorithm}[H]
\caption{Diffusion-based Quality Estimator}\label{alg:algorithm}
\KwIn{
    $X_{HQ}, X_{LQ}$: Two groups of train images of higher and lower visual quality \\
    $X$: Test images, $\left| X\right| = N$ \\
    $\epsilon_\theta$: Pretrained text-to-image generative model \\
    $\mathcal{P}$: Predefined prompt with tokens $\{w_1, ..., w_M\}$ \\
    $L$: Number of cross-attention layers \\
    $K$: Number of top discriminative features \\
    $t$: Timestep for activation extraction
}
\KwOut{Quality scores $\mathbf{f} \in \mathbb{R}^{N}$}

\SetAlgoLined

1. \textbf{Extract activations:} \\
\For{each image $x \in \mathbb{R}^{h \times w}$ in $X_{HQ} \cup X_{LQ} \cup X$}{
    Save cross-attn maps \{$A^{(x)}_{l, m} \in \mathbb{R}^{h_{l} \times w_{l}}\}_{\substack{l=1 
\dots L \\ m=1 \dots M}}$ during noise prediction via $\epsilon_\theta{(x, \mathcal{P}, t)}$ \\
    Compute spatial activation norms: \\
    \quad $N^{(x)}_{l,m} = \|A^{(x)}_{l,m,:,:}\|_2 \quad \forall l \in \{1,...,L\}, m \in \{1,...,M\}$
}

2. \textbf{Find (layer, token) pairs with most discriminative features:} \\
\For{each $(l, m)$ pair}{
    \quad $s_{l,m} \leftarrow 0$ \\
    \quad \For{each $(x_{HQ} \in X_{HQ}, x_{LQ} \in X_{LQ})$ pair}{
    \quad\quad Compute separation score: \\
    \quad\quad $s_{l,m} \mathrel{+}= \mathbb{I}{[ N^{(x_{HQ})}_{l,m} > N^{(x_{LQ})}_{l,m}]}$ \\
    \quad }
}
Select top-$K$ $(l,m)$ pairs with highest $s_{l,m}$: $\mathcal{K} = \{(l_1,m_1), ..., (l_K,m_K)\}$

3. \textbf{Compute scores:} \\
\For{each image $x \in X$}{
    $\mathbf{f}_x = \sum\limits_{(l,m) \in \mathcal{K}} N^{(x)}_{l,m}$
}

\Return{Quality scores $\mathbf{f}$}
\end{algorithm}
\end{minipage}
\vspace{-15px}
\end{figure}

The objective here is to find a subset of images with a rare combination of visual characteristics such as high aesthetic quality, optimal color balance, and substantial image complexity that are hypothesized to maximize SFT quality.
Existing IQA and IAA models often struggle to holistically capture this specific blend of attributes crucial for SFT.

Our hypothesis is that a pre-trained diffusion model, through its learned representations, inherently encodes these desired characteristics, particularly within its cross-attention mechanisms which mediate text-image alignment during generation.
To leverage this, we developed a scoring function based on cross-attention activations.
We utilize a long, multi-keyword prompt designed to evoke the target visual qualities (e.g., including terms like ``high quality``, ``artistic``, ``aesthetic``, ``complex``).
For each image, we extract cross-attention activation norms  corresponding to these keywords.
To identify the most discriminative activations, we manually scored a calibration set of 1,000 images based on the aforementioned SFT-desirable criteria, forming ``higher-quality`` and ``lower-quality`` groups. 
We then identified the top-$K$ activation indices that best separated these two groups.
The final score for any given image is an aggregation (summation) of its activation norms at these top-$K$ indices (details in Algorithm \ref{alg:algorithm}).
A detailed discussion of methodological details, including prompt engineering and choice of inference timestep $t$, is provided in Appendix \ref{appendix:dataset_collection_details}.

Using this diffusion-based scoring function, we evaluated all $N\approx$ 300 million images from Stage~3 and selected the top-$n$ samples.
The SFT dataset size ($n$) is a critical hyperparameter. 
Through ablation studies (detailed in Section \ref{subsec:dataset_size_ablation}), we determined that $n =$ 3,350 provides best model quality improvements with no observable loss of generative diversity.

\vspace{-5px}
\paragraph{Final Re-captioning and the Alchemist Dataset.}

The 3,350 images curated by our pipeline, though visually exceptional, retained their original, often noisy, web captions. 
Effective supervised fine-tuning (SFT) necessitates appropriate textual guidance.
Our preliminary studies highlighted the importance of caption style, finding that captions resembling moderately descriptive user-like prompts rather than exhaustively detailed ones achieve better SFT results.
Therefore, we re-captioned the entire set using a proprietary image captioning model tuned to produce such user-centric descriptions.
This re-captioning ensured consistent and relevant textual pairings.
The resulting Alchemist dataset consists of these 3,350 refined image-text pairs, used for subsequent analysis and SFT.

\vspace{-10px}
\section{Experiments}
\vspace{-8px}
\label{sec:experiments}

We empirically evaluate the effectiveness of \textbf{Alchemist} as an SFT dataset for open-source Stable Diffusion (SD) models. 
Our goal is to verify whether a compact, highly curated dataset like Alchemist can significantly boost image generation quality and outperform LAION-Aesthetics v2 \cite{laion-aesthetics-v2} as a standard publicly available SFT alternative.
Below we discuss experimental setup and present results of fine-tuning with Alchemist in terms of human-perceived generation quality and common automated metrics.
\vspace{-5px}
\subsection{Experimental Setup}\label{subsec:experimental_setup}

\paragraph{Models and Datasets.}
We evaluate our proposed methodology across five widely-used pre-trained text-to-image models based on Stable Diffusion: SD1.5\footnote{\url{https://huggingface.co/stable-diffusion-v1-5/stable-diffusion-v1-5}}, SD2.1\footnote{\url{https://huggingface.co/stabilityai/stable-diffusion-2-1}}, SDXL1.0\footnote{\url{https://huggingface.co/stabilityai/stable-diffusion-xl-base-1.0}}, SD3.5 Medium\footnote{\url{https://huggingface.co/stabilityai/stable-diffusion-3.5-medium}}, and SD3.5 Large\footnote{\url{https://huggingface.co/stabilityai/stable-diffusion-3.5-large}} \cite{rombach2022high, podell2023sdxl, esser2024scaling}. 
For each base model, we utilize the official checkpoints and Diffusers-based \cite{von-platen-etal-2022-diffusers} publicly available fine-tuning code to establish three comparison points:

\begin{itemize}
    \item \textbf{Baseline:} The original official model weights;
    \item \textbf{Alchemist-tuned:} The baseline model fine-tuned on our proposed Alchemist dataset (comprising 3,350 samples);
    \item \textbf{LAION-tuned:} The baseline model fine-tuned on a size-matched subset (3,350 samples) drawn from the LAION-Aesthetics v2 dataset \cite{laion-aesthetics-v2}, specifically selecting samples with aesthetic scores $>=$ 6.5. 
    This serves as a control to assess the effectiveness of Alchemist compared to a standard, high-aesthetics filtered dataset of equivalent size.
    We additionally ablate dataset size for LAION in Appendix \ref{appendix:laion_size}.
\end{itemize}

\paragraph{Fine-Tuning and Hyperparameter Selection.}
We employed a full fine-tuning approach, updating all parameters of the base models. 
To identify optimal settings for each (model, dataset) combination, we conducted a grid search over key hyperparameters, including learning rate, EMA momentum, and the number of training steps.
The specific search ranges and the final selected hyperparameters for each configuration are detailed in Appendix \ref{appendix:sweep}.
Checkpoint selection and early stopping decisions during this tuning process were guided by performance on a dedicated validation set.
This validation set consisted of 128 prompts selected from the PartiPrompts benchmark \cite{yu2022scaling}, following the methodology employed in SD3 \cite{esser2024scaling}.

\paragraph{Test Set for Final Evaluation.}
The final performance assessment of the best checkpoints selected via the validation process was conducted on a separate, unseen test set. 
This test set comprised 500 distinct prompts also drawn from PartiPrompts \cite{yu2022scaling}, ensuring no overlap with the prompts used during validation or hyperparameter tuning.

\subsection{Evaluation Protocol}\label{subsec:eval_protocol}

\paragraph{Human Side-by-Side Evaluation}
Our primary method for evaluating model performance relies on human perception via side-by-side (SbS) comparisons.  
For each comparison pair (e.g., Alchemist-tuned vs. Baseline), we generated images using prompts from the validation or test sets; detailed parameters are provided in Appendix \ref{appendix:inference_parameters}).
Three expert annotators were independently presented with the generated images. 
Annotators evaluated the pairs based on four criteria:
\begin{itemize}
    \item \textbf{Image-Text Relevance:} Accuracy of the image content relative to the text prompt;
    \item \textbf{Aesthetic Quality:} Overall visual appeal, including composition and style;
    \item \textbf{Image Complexity:} Richness of detail and content within the scene;
    \item \textbf{Fidelity:} Presence and severity of defects, artifacts, distortions, or undesirable elements.
\end{itemize}

For each criterion, annotators selected the preferred image, with the option of indicating a tie. 
The final outcome for a given prompt and criterion was determined by majority vote among the three annotators. 
We assess the statistical significance of the aggregate win rates using a two-sided binomial test ($p < 0.05$).
Details regarding the SbS interface and instructions are provided in Appendix \ref{appendix:sbs}.

\paragraph{Automated Metrics}
To complement human judgments, we report established automated metrics. 
These include \textbf{FD-DINOv2}, which calculates the Fr\'echet Distance \cite{heusel2017gans} using DINOv2 \cite{oquab2023dinov2} features, and \textbf{CLIP Score} \cite{hessel2021clipscore}, based on ViT-L/14 \cite{dosovitskiy2020image} image-text similarity. 
Additionally, we employ learned human preference predictors: \textbf{ImageReward} (IR) \cite{xu2023imagereward} and \textbf{HPS-v2} \cite{wu2023human}. 
All automated metrics were computed on the standard MJHQ-30K dataset \cite{li2024playground}. 

\subsection{Results}
\label{subsec:results}

\begin{table}[tbp]
    \centering
    \resizebox{0.99\textwidth}{!}{
        \begin{tabular}{lccccccccc}
  \toprule
  \thead[c]{\multirow{2}{*}{Model}} 
  & \multicolumn{4}{c}{\textbf{Side-by-Side Win Rate}} 
  & \multicolumn{4}{c@{}}{\textbf{Automatic Metrics} ($\Delta$)} \\
  \cmidrule(lr){2-5} 
  \cmidrule(lr){6-9}
  & Rel.$\uparrow$  
  & Aes. $\uparrow$
  & Comp. $\uparrow$ 
  & Fidel. $\uparrow$ 
  & FD$_{\text{DINOv2}}$ $\downarrow$ 
  & CLIP $\uparrow$ 
  & IR $\uparrow$ 
  & HPS-v2 $\uparrow$ \\
  \midrule
  \textbf{SD1.5-Alchemist}
  & & & & & 129.8 & 0.277 & \textbf{0.38} & \textbf{0.270} \\
  \quad vs baseline 
  & \neutral{0.53} 
  & \good{0.64} 
  & \good{0.78} 
  & \neutral{0.47} 
  & 131.5  
  & 0.279  
  & 0.02 
  & 0.243 
  \\
  \quad vs LAION-tuned 
  & \neutral{0.47} 
  & \good{0.60} 
  & \good{0.73} 
  & \bad{0.45} 
  & \textbf{112.1} 
  & \textbf{0.286} 
  & 0.32 
  & 0.260 
  \\ 
  \cmidrule(lr){1-1} \cmidrule(lr){2-5} \cmidrule(lr){6-9}
  \textbf{SD2.1-Alchemist}
  & & & & & \textbf{95.6} & 0.281 & 0.62 & \textbf{0.282} \\
  \quad vs baseline 
  & \good{0.57} 
  & \good{0.69} 
  & \good{0.81} 
  & \good{0.56} 
  & 129.3 
  & 0.276 
  & 0.18 
  & 0.253 
  \\
  \quad vs LAION-tuned 
  & \neutral{0.49} 
  & \good{0.56}
  & \good{0.72}
  & \neutral{0.52}
  & 112.4
  & \textbf{0.287} 
  & \textbf{0.65} 
  & 0.278 
  \\ 
  \cmidrule(lr){1-1} \cmidrule(lr){2-5} \cmidrule(lr){6-9}
  \textbf{SDXL-Alchemist}
  & & & & & 97.4 & 0.286 & 0.76 & \textbf{0.292}  \\
  \quad vs baseline 
  & \neutral{0.52}
  & \good{0.61}
  & \good{0.78}
  & \neutral{0.51}
  & \textbf{73.4} 
  & 0.293 
  & 0.71 
  & 0.283 
  \\
  \quad vs LAION-tuned 
  & \neutral{0.49}
  & \good{0.58}
  & \good{0.78}
  & \good{0.57}
  & 108.9 
  & \textbf{0.294} 
  & \textbf{0.81} 
  & 0.291 
  \\
  \cmidrule(lr){1-1} \cmidrule(lr){2-5} \cmidrule(lr){6-9}
  \textbf{SD3.5M-Alchemist}
  & & & & & \textbf{76.2} & 0.286 & \textbf{1.07} & \textbf{0.295} \\
  \quad vs baseline & \neutral{0.51}
  & \good{0.57}
  & \good{0.67}
  & \neutral{0.50}
  & 81.4
  & \textbf{0.287}
  & 0.97
  & 0.292 \\
  \quad vs LAION-tuned 
  & \neutral{0.48}
  & \good{0.58}
  & \good{0.73}
  & \neutral{0.49}
  & 87.9
  & 0.286
  & 0.87 
  & 0.274 
  \\
  \cmidrule(lr){1-1} \cmidrule(lr){2-5} \cmidrule(lr){6-9}
  \textbf{SD3.5L-Alchemist}
  & & & & & \textbf{80.9} & 0.287 & \textbf{1.12} & \textbf{0.299} \\
  \quad vs baseline & \neutral{0.49}
  & \good{0.62}
  & \good{0.72}
  & \bad{0.41}
  & 91.4 & 0.286 & 1.01 & 0.298  \\
  \quad vs LAION-tuned 
  & \neutral{0.47}
  & \good{0.57}
  & \good{0.76}
  & \neutral{0.55}
  & 91.1
  & \textbf{0.297}
  & 1.10
  & 0.294
  \\
  \bottomrule
\end{tabular}

    }
    \vspace{4pt}
    \caption{Comparison of Alchemist-tuned models, baselines, and LAION-Aesthetics-tuned models. 
    The table reports human win rates (by aspect) w.r.t. Alchemist-tuned models and automated metrics values for each model variant.
    \good{Green} indicates statistically significant improvement ($p < 0.05$), \neutral{gray} no statistically significant change, and \bad{red} a statistically significant decline. 
    For automated metrics \textbf{bold} means the best value among three model variants.}
    \label{tab:model-comparison}
    \vspace{-7px}
\end{table}

\paragraph{Human Evaluation Results.}
The results from human side-by-side (SbS) evaluations demonstrate how fine-tuning impacts the four specified assessment criteria. 
Regarding \textbf{Image-Text Relevance}, fine-tuning with Alchemist did not yield statistically significant differences compared to either the baseline or the LAION-tuned models across most tested architectures ($p > 0.05$).
This indicates that the improvements observed in other aspects do not compromise prompt fidelity.

Conversely, Alchemist fine-tuning demonstrated substantial and statistically significant improvements in both \textbf{Aesthetic Quality} and \textbf{Image Complexity}. 
Compared to the respective baseline models, Alchemist-tuned versions achieved human preference win rates up to 20\% higher.
Furthermore, Alchemist consistently outperformed the size-matched LAION-Aesthetics-tuned variants on these two criteria, with win rate advantages ranging from +12\% to +20\% across the different base models.

In terms of \textbf{Fidelity}, the results were mixed.
While many models showed no significant change, fine-tuning with Alchemist led to a marginal but statistically significant decrease in perceived fidelity for certain architectures (average win rate decrease of approximately 5\% against baseline in those cases).
We hypothesize this may represent a tradeoff associated with generating more complex and detailed images, a point further discussed in Sections \ref{sec:discussion} and \ref{sec:limitations}.

\vspace{-5px}
\paragraph{Qualitative Analysis}

\begin{figure}[htbp]
    \centering
    \begin{subfigure}[b]{0.49\textwidth}
        \includegraphics[width=\textwidth]{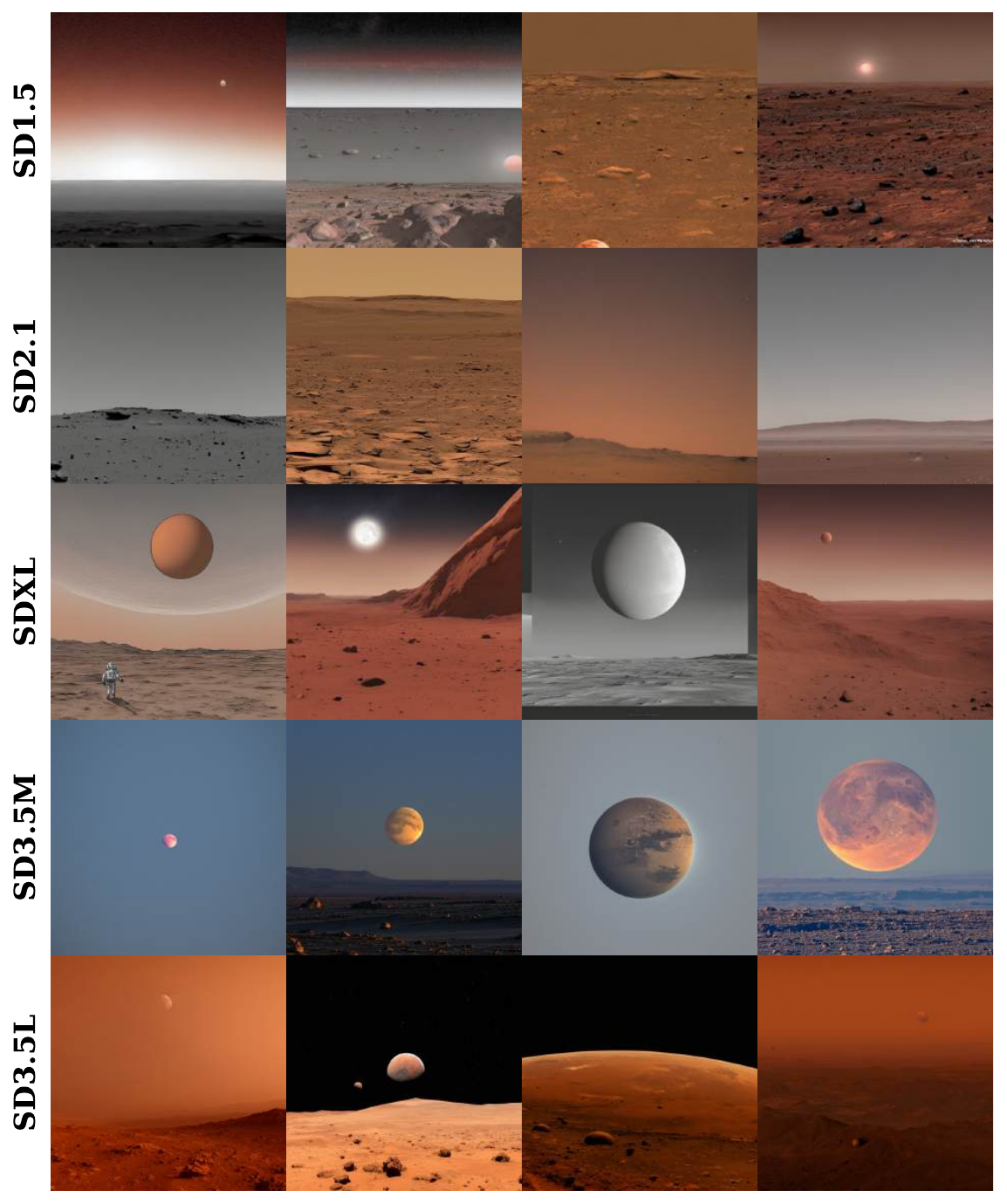}
        \label{fig:before_tune_all}
        \vspace{-8px}
        \caption{Baseline.}
    \end{subfigure}
    \hfill 
    \begin{subfigure}[b]{0.473\textwidth}
        \includegraphics[width=\textwidth]{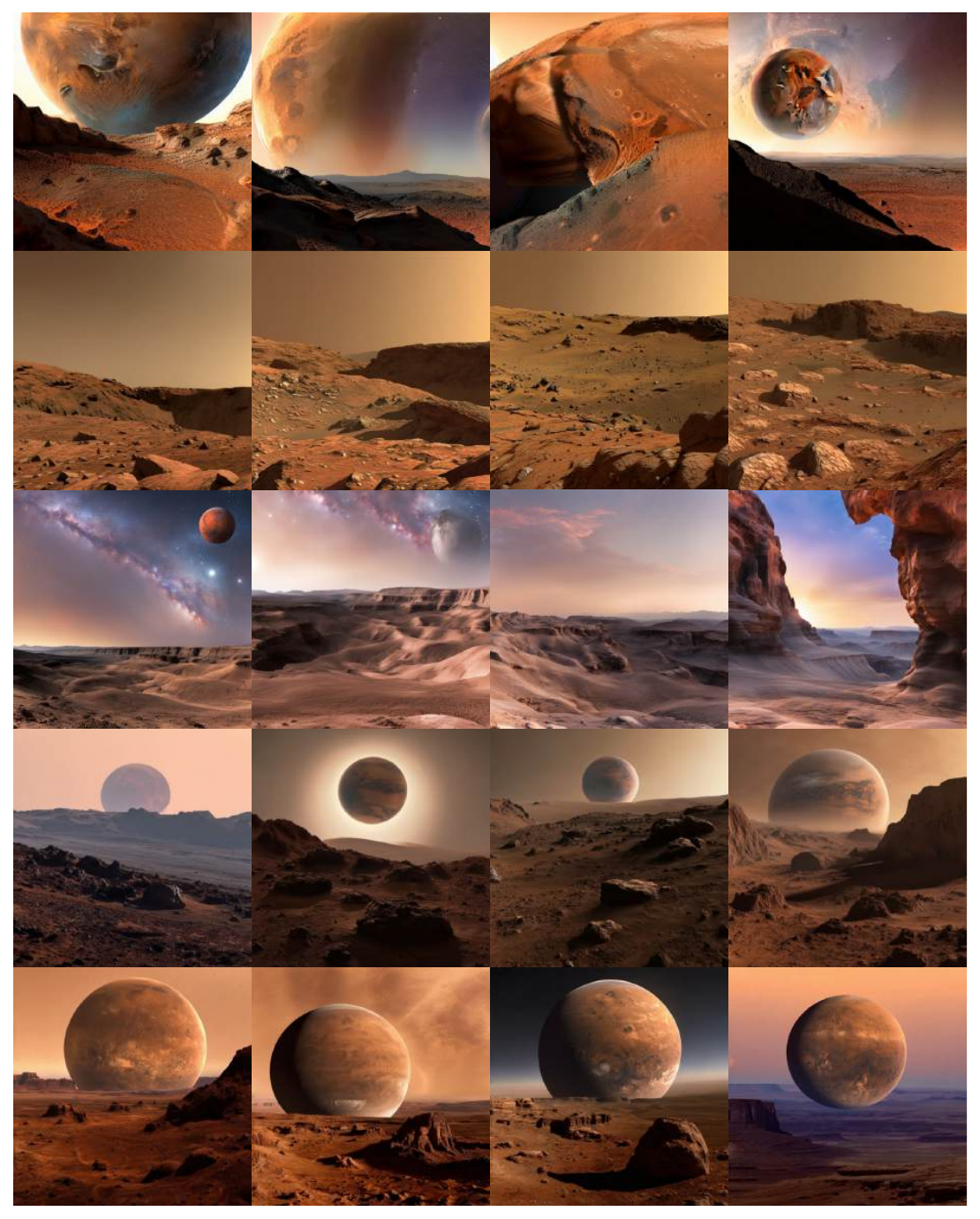}
        \label{fig:after_tune_all}
        \vspace{-8px}
        \caption{Alchemist-tuned.}
    \end{subfigure}
    \caption{Examples of images generated by five Stable Diffusion models for the prompt ``\textit{Mars rises on the horizon.}'' \textbf{before} and \textbf{after} tuning on Alchemist.}
    \label{figure:sft_qualitative_results}
    \vspace{-7px}
\end{figure}

To visually complement these quantitative assessments and human judgments, Figure \ref{figure:sft_qualitative_results} presents qualitative examples of images generated by several models, showcasing outputs from their baseline versions alongside those after fine-tuning with Alchemist.
These visual comparisons directly illustrate the enhancements in aesthetic appeal, detail, and overall image complexity reported above.
The examples also suggest that fine-tuning with Alchemist does not lead to a noticeable decline in the diversity of content or stylistic range generated by the models. 
A more extensive collection of qualitative results, including additional model comparisons and prompt examples, is provided in Appendix \ref{appendix:more_vis}.

\paragraph{Automated Metric Results.}
These findings from human evaluations and qualitative analysis are further confirmed by automated metrics. 
Improvements in FD-DINOv2, CLIP Score, and the learned preference scores (ImageReward, HPS-v2) were observed for most models after fine-tuning with Alchemist, particularly when compared to the untuned baselines (see Table \ref{tab:model-comparison} for detailed results).
The comparison against LAION-Aesthitics-tuned models on these metrics also generally favored the Alchemist variants, supporting the conclusions drawn from human preferences.

\subsection{Dataset Size Ablation}
\label{subsec:dataset_size_ablation}

To assess the impact of strict filtering, we created two larger Alchemist variants (approx. 7k and 19k samples) by relaxing the selection threshold of our diffusion-based quality estimator. 
These datasets inherently contained samples with lower diffusion-guided quality scores than the original 3,350-sample Alchemist. 
We then fine-tuned all five base models on these 7k and 19k variants.

As summarized in Table \ref{tab:alchemist-size-ablation}, fine-tuning on both larger datasets yielded consistently inferior performance across all models compared to the compact 3,350-sample Alchemist.
An additional, dedicated hyperparameter sweep for the 7k and 19k datasets (Appendix \ref{appendix:additional_ablations}) confirmed this finding, as no configuration achieved quality comparable to that of the original Alchemist.
These results underscore that exceptional sample quality curated by strict, diffusion-guided filtering is more critical for maximizing SFT efficacy than sheer dataset volume. 

\section{Discussion}
\label{sec:discussion}


Fine-tuning with Alchemist substantially enhances aesthetic quality and image complexity across diverse Stable Diffusion models, highlighting the power of targeted SFT with compact, high-impact datasets. 
Our findings, however, also prompt further discussion.

\begin{figure}[htbp]
    \centering
    \begin{minipage}{0.47\textwidth}
        \resizebox{1.05\textwidth}{!}{
        \begin{tabular}{lccccc}
  \toprule
  \thead[c]{\multirow{2}{*}{Model}} 
  & \multicolumn{4}{c}{\textbf{Side-by-Side Win Rate}} 
   \\
  \cmidrule(lr){2-5} 
  & Rel.$\uparrow$  
  & Aes. $\uparrow$
  & Comp. $\uparrow$ 
  & Fidel. $\uparrow$ \\
  \midrule
  \textbf{SD1.5-Alchemist-3k}
  & & & & & \\
  \quad vs Alchemist-7k 
  & \bad{0.44} 
  & \good{0.62} 
  &  \good{0.64} 
  &  \neutral{0.47} 
  \\
  \quad vs Alchemist-19k 
  & \bad{0.43} 
  & \good{0.62} 
  & \good{0.67} 
  & \neutral{0.49} 
  \\ 
  \cmidrule(lr){1-1} \cmidrule(lr){2-5}
  \textbf{SD2.1-Alchemist-3k}
  & & & & & \\
  \quad vs Alchemist-7k 
  &  \neutral{0.45} 
  &  \good{0.61} 
  &  \good{0.62} 
  &  \neutral{0.55} 
  \\
  \quad vs Alchemist-19k 
  &  \neutral{0.46} 
  &  \good{0.60}
  &  \good{0.76}
  &  \neutral{0.53}
  \\ 
  \cmidrule(lr){1-1} \cmidrule(lr){2-5}
  \textbf{SDXL-Alchemist-3k}
  & & & & & \\
  \quad vs Alchemist-7k 
  &  \neutral{0.49}
  &  \good{0.61}
  &  \good{0.66}
  &  \good{0.57}
  \\
  \quad vs Alchemist-19k 
  &  \neutral{0.48}
  &  \good{0.65}
  &  \good{0.73}
  &  \neutral{0.53}
  \\
  \cmidrule(lr){1-1} \cmidrule(lr){2-5}
  \textbf{SD3.5M-Alchemist-3k}
  & & & & & \\
  \quad vs Alchemist-7k  
  &  \neutral{0.48}
  &  \good{0.61}
  &  \good{0.58}
  &  \neutral{0.53}
  \\
  \quad vs Alchemist-19k  
  &  \neutral{0.50}
  &  \good{0.75}
  &  \good{0.81}
  &  \good{0.58}
  \\
  \cmidrule(lr){1-1} \cmidrule(lr){2-5}
  \textbf{SD3.5L-Alchemist-3k}
  & & & & & \\
  \quad vs Alchemist-7k 
  &   \neutral{0.54}
  &  \good{0.52}
  &  \good{0.47}
  &  \neutral{0.55}
 \\
  \quad vs Alchemist-19k
  &  \neutral{0.52}
  &  \good{0.68}
  &  \good{0.70}
  &  \good{0.57}
  \\
  \bottomrule
\end{tabular}

    }
    \vspace{4pt}
    \caption{Comparison of models fine-tuned on Alchemist variants of different sizes.
    The table reports human win rates (by aspect) of Alchemist-3k-tuned models against models tuned on 7k and 19k variants of Alchemist.
    \good{Green} indicates statistically significant improvement ($p < 0.05$), \neutral{gray} no significant change, and \bad{red} a statistically significant decline.}
    \label{tab:alchemist-size-ablation}

    \end{minipage}
    \hfill
    \begin{minipage}{0.41\textwidth}
        \centering
        \includegraphics[width=\textwidth]{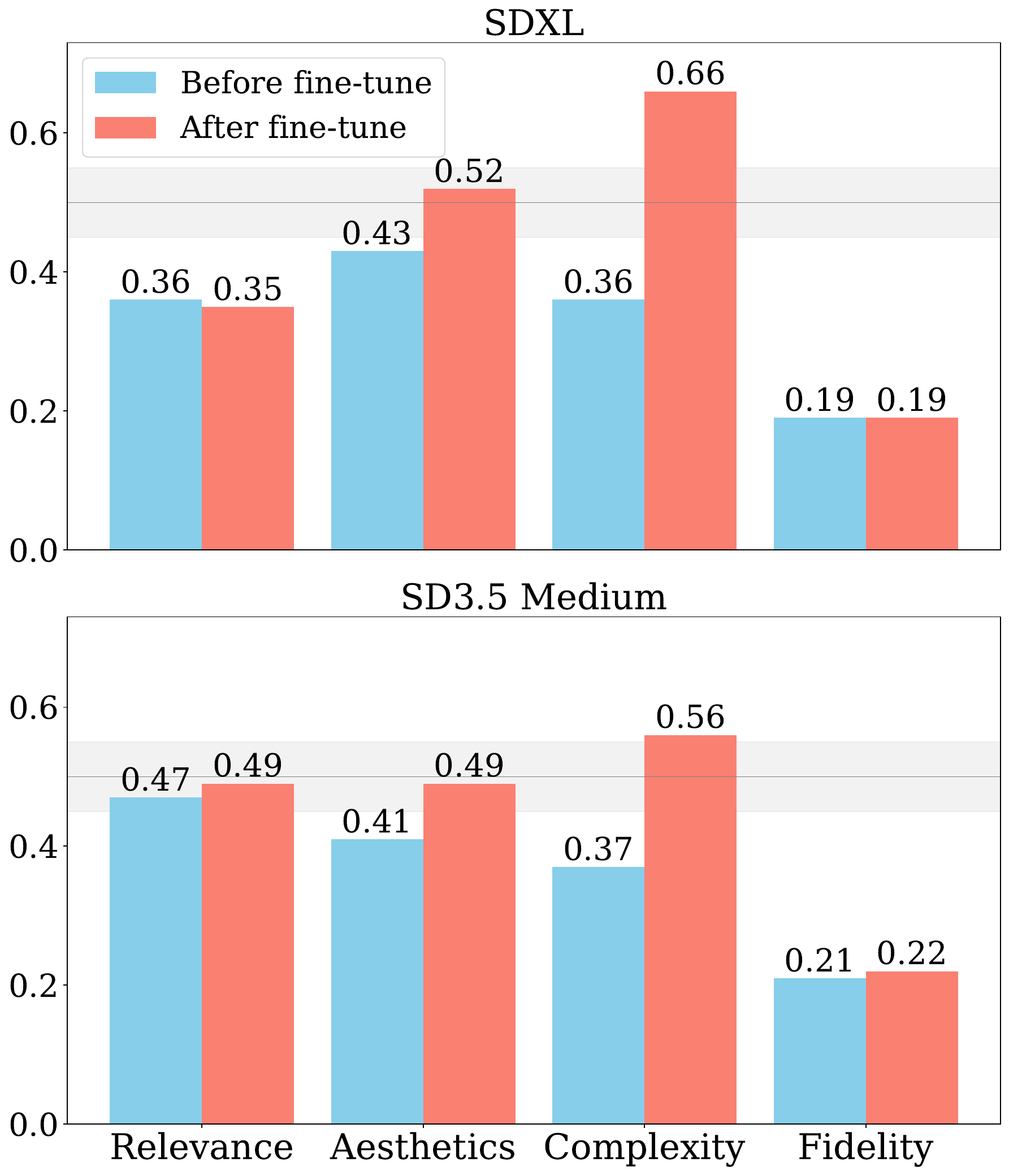}
        \caption{ Results of SbS comparison of SDXL, SD3.5 Medium before and after fine-tuning versus FLUX. Grey shaded region shows the interval of statistical insignificance.}
        \label{fig:vs_flux_main}
    \end{minipage}
\end{figure}

We observe that Alchemist fine-tuning yields varied improvements and trade-offs across models. 
Notably, later architectures like SD3.5 showed a slight decrease in fidelity, a trend less apparent in earlier models. 
This difference likely stems from the base models' histories: newer models may have already incorporated some variation of fine-tuning after their initial pre-training. 
Consequently, our general-purpose SFT with Alchemist, while beneficial, might introduce characteristics that subtly conflict with these existing, highly specific optimizations, leading to the observed fidelity trade-off. 
Earlier models, with less such prior refinement, may more readily absorb Alchemist's broad quality enhancements.

We also observe an inherent link between increased image complexity and a potential drop in fidelity.
Guiding models towards richer scenes, a strength Alchemist confers, inherently provides more opportunities for minor artifacts. 
This suggests that achieving high complexity and maximal fidelity may necessitate techniques beyond general SFT.

Furthermore, our results confirm this SFT approach minimally impacts image-text relevance. 
This aspect seemingly depends more on model architecture, initial pre-training data, and dedicated alignment methods, rather than fine-tuning primarily focused on visual style.


%

Ultimately, Alchemist's quality improvements effectively bridge the performance gap between traditional SD models and cutting-edge solutions. 
Figure \ref{fig:vs_flux_main} reveals that Alchemist-tuned SDXL and SD3.5 Medium exhibit aesthetic quality and image complexity comparable to leading models like FLUX.1-dev \cite{flux2024} despite having 4 times less parameters. 
This underscores that data-efficient SFT on well-pre-trained foundations remains a viable path to significant quality advancements.

\section{Limitations}
\label{sec:limitations}


While Alchemist fine-tuning significantly enhances image aesthetics and complexity (Section \ref{subsec:results}), two primary limitations warrant acknowledgment. 
Firstly, this pursuit of visual richness can introduce a marginal decrease in perceived fidelity for some models, a trade-off more pronounced in highly optimized later architectures (e.g., SDXL, SD3.5) than in earlier ones (e.g., SD1.5, SD2.1) which showed clearer net quality gains without substantial defect increases.
This suggests that pushing already high-performing models towards greater complexity via SFT may inherently surface minor imperfections.
Secondly, our approach did not yield significant improvements in image-text relevance.
This aspect appears to be more dependent on factors like model architecture, initial pre-training, and dedicated alignment techniques, rather than the visual quality-focused SFT employed here.
Despite these points, Alchemist effectively achieves its primary goal of elevating key visual qualities in text-to-image models using a compact, targeted dataset.

\section{Conclusion}
\label{sec:conclusions}

This work introduced Alchemist, a compact (3,350 samples) supervised fine-tuning (SFT) dataset, and its novel creation methodology leveraging a pre-trained diffusion model as a key quality estimator, followed by re-captioning with moderately descriptive, user-like prompts. 
Extensive experiments across five Stable Diffusion models demonstrated Alchemist's effectiveness in significantly boosting aesthetic quality and image complexity, outperforming baselines and a size-matched LAION-Aesthetics SFT.
While image-text relevance remained largely unaffected and a minor complexity-fidelity trade-off emerged for highly optimized models, ablation studies underscored the crucial role of our strict filtering and compact dataset size for achieving superior SFT outcomes. 
Our principled, data-efficient approach and the public release of the Alchemist dataset and fine-tuned model weights offer valuable resources and insights for advancing text-to-image generation through high-quality SFT.

\bibliographystyle{plain}  
\bibliography{references}

\newpage
\appendix

\section*{Appendices}

Here we show additional results on generation of non-square images (Appnedix \ref{appendix:non_square}), detail dataset collection procedure (Appendix \ref{appendix:dataset_collection_details}), provide additional ablations of data filtering and LAION-Aesthetics size selection (Appendix \ref{appendix:additional_ablations}), detail our experimental and inference settings (Appendices \ref{appendix:experimental_setting} and \ref{appendix:inference_parameters}), and describe and provide examples of human evaluation (Appendix \ref{appendix:sbs}). 
In conclusion we discuss broader impact (Appendix \ref{appendix:broader_impact}) and provide addisional visualizations for qualitative assessment (Appendix \ref{appendix:more_vis}).

\section{Non-square Aspect Ratio Generation}
\label{appendix:non_square}

In the early era of diffusion-based text-to-image generation, models such as SD1.5 and SD2.1 were trained exclusively on square images, limiting their ability to generate images with different aspect ratios. 
As diffusion technology advanced, the concept of bucketed training was introduced \cite{novelai2023improvements}.
This approach organizes training batches by resolution, where each batch contains images of identical resoluitons, but the image size varies between training iterations. 
This methodology enabled models to generate images across diverse aspect ratios.

The Alchemist dataset comprises images with varying aspect ratios, facilitating bucketed fine-tuning. 
This training approach ensures the model can produce high-quality images beyond the traditional square format.

In our experiments with SDXL, SD3.5 Medium, and SD3.5 Large, we implemented bucketing for the latent representations, varying the latent resolution across batches. In addition to evaluating square image generation, we present side-by-side (SbS) comparisons between Alchemist-tuned versions of SDXL, SD3.5 Medium, and SD3.5 Large against their original counterparts, which were inherently designed to support multi-aspect ratio image generation. 

\begin{table}[htbp]
    \centering
    {
        \begin{tabular}{lccccc}
  \toprule
  \thead[c]{\multirow{2}{*}{Model}} 
  & \multicolumn{4}{c}{\textbf{Side-by-Side Win Rate}} 
   \\
  \cmidrule(lr){2-5} 
  & Rel.$\uparrow$  
  & Aes. $\uparrow$
  & Comp. $\uparrow$ 
  & Fidel. $\uparrow$ \\
  \midrule
  \textbf{SDXL-Alchemist}$_{[h,w]}$
  & & & & & \\
  \quad vs baseline$_{[1280,768]}$
  & \neutral{0.53} 
  & \good{0.62} 
  &  \good{0.78} 
  & \neutral{0.49} 
  \\
  \quad vs baseline$_{[896,1152]}$
  & \neutral{0.48} 
  & \good{0.63} 
  &  \good{0.83} 
  & \neutral{0.49} 
  \\

  \cmidrule(lr){1-1} \cmidrule(lr){2-5}
  \textbf{SD3.5M-Alchemist}$_{[h,w]}$
  & & & & & \\
  \quad vs baseline$_{[1280,768]}$
  &  \neutral{0.52} 
  &  \good{0.55} 
  &  \good{0.65} 
  &  \neutral{0.50} 
  \\
  \quad vs baseline$_{[896,1152]}$
  & \neutral{0.50} 
  & \good{0.60} 
  & \good{0.70} 
  & \neutral{0.51} 
  \\

  \cmidrule(lr){1-1} \cmidrule(lr){2-5}
  \textbf{SD3.5L-Alchemist}$_{[h,w]}$
  & & & & & \\
  \quad vs baseline$_{[1280,768]}$
  & \neutral{0.51} 
  & \good{0.60} 
  &  \good{0.71} 
  & \bad{0.40} 
  \\
  \quad vs baseline$_{[896,1152]}$
  & \neutral{0.51} 
  & \good{0.67} 
  & \good{0.72} 
  & \bad{0.40} 
  \\

  \bottomrule
\end{tabular}
    }
    \vspace{4pt}
    \caption{
    We evaluated the ability of Alchemist-tuned models and baseline models to generate images with non-square aspect ratios. 
    For each model, we produced images of resolution \([h, w]\), where the exact values of \(h\) and \(w\) are specified in the baseline subscripts. 
    The table reports human win rates (by aspect) w.r.t. Alchemist-tuned models.
    \good{Green} indicates statistically significant improvement ($p < 0.05$), \neutral{gray} no statistically significant change, and \bad{red} a statistically significant decline.}
    \label{tab:non-square}
\end{table}

The side-by-side (SbS) comparison results align with those in Table \ref{tab:model-comparison}: fine-tuning enhances the aesthetic quality and complexity of generated images without sacrificing text coherence, though for SD3.5 Large it occasionally introduces more artifacts due to increased detail. 
This confirms that the dataset enables generation of images of various aspect ratios without compromising their quality compared to the baseline square size.


    

\section{Dataset Collection Details}
\label{appendix:dataset_collection_details}

\subsection{Timestep Selection}
\label{app_sebsec:timestep_selection}

The timestep \( t \in [0.0, 1.0] \) in the input in Algorithm \ref{alg:algorithm} is a crucial parameter of our approach. 
When \( t \) approaches $0.0$, the generated image is almost fully formed, and the influence of the text prompt diminishes significantly.
Conversely, as \( t \) approaches $1.0$, the activations become dominated by noise and lose interpretability. 
Through empirical analysis, we identified $t = 0.25$ as an optimal balance point and employed this value across all binary classifiers.


\subsection{Diffusion-based Estimator Prompt}
\label{app_sebsec:prompt_keyword}

Another critical element of Algorithm \ref{alg:algorithm} is its input text prompt $\mathcal{P}$. We define it as follows:

\begin{tcolorbox}
\textit{"complex. detailed. simple. bokeh effect. abstract. photorealistic. artistic. 
stylized. aesthetic. cinematic. instagram filters. color correction. midjourney. ugly. distorted. blurry. rendering. AI-generated. synthetic. high quality. low quality. pixelated. low illumination."}
\end{tcolorbox}

This prompt formulation integrates both empirical findings and theoretical principles of visual appeal, specifically targeting perceptual factors that influence human judgments of image quality. 
The template incorporates descriptors that capture both desirable and undesirable attributes across key visual dimensions:

\begin{enumerate}
    \item \textbf{Image complexity.}  Our experimental analysis revealed that images with minimal visual complexity (e.g., images with monochrome backgrounds or reduced detail density) contributed negligibly to model generation quality and are being overshadowed by more intricate, information-rich counterparts. 
    Furthermore, while the inclusion of images featuring bokeh effects demonstrated a stabilizing influence on training dynamics, we observed a corresponding degradation in overall model performance. 
    Consequently, our final curation pipeline excluded both minimalistic imagery and samples exhibiting excessive bokeh distortion.

    \item \textbf{Art.} Artistic images and real life photos inherently differ in their visual characteristics and require separate processing pipelines. 
    Photographic quality relies on objective technical parameters that are more suited for measurements, whereas artistic quality depends on subjective stylistic choices and are often out-of-domain for the most of classifiers. 
    For these reasons we focus on incorporating such feature in our prompt.

    \item \textbf{Aesthetic and Color correction.} We aim to estimate the aesthetic quality of images by learning discriminative features associated with coherent color palette, sharp focus on key subjects, satisfying photo composition rules and other properties of aesthetically compelling images from those that are commonly produced by amateur photography. 
    A critical subtask in computational aesthetic evaluation involves assessing color fidelity, as a significant portion of consumer-grade photographs exhibit improper white balance, inaccurate saturation, or unnatural tonal distributions due to uncalibrated capture devices and lack of skill. 
    This aspect specifically identifies images with professional-grade color correction characterized by balanced neutral tones, highlight-to-shadow transitions, proper color palette and saturation.

    \item \textbf{Compression and noise.} Beyond aesthetic considerations, technical image quality presents a  challenge for generative model training. 
    Degradation categories, such as compression artifacts from JPEG and WebP formats, sensor-level noise and optical aberrations, affect high-frequency features learning that results with increase in image generation artifacts.
\end{enumerate}

\section{Additional Ablations}
\label{appendix:additional_ablations}
\subsection{Filtration Approach}
\label{appendix:filtration}

This subsection examines the necessity of the diffusion-based estimator in our filtration pipeline. 
To evaluate its importance, we removed this component and implemented a more rigorous filtering process using TOPIQ-IAA \cite{chen2024topiq} and classifiers trained on TAD-66k \cite{he2022_TAD}, KonIQ-10k \cite{koniq10k} and IC-9600 \cite{ic9600feng}.
This approach maintained the same sample size as Alchemist while selecting for high aesthetic quality and substantial complexity. 
All other steps in our filtration pipeline remained unchanged.

We adhere to the same image appeal considerations detailed in Appendix~\ref{appendix:dataset_collection_details}.
Our pipeline begins with complexity filtering using the IC-9600 classifier, where we apply a lower threshold to exclude monochromatic or overly simplistic images. 

Next, we employ aesthetic and image quality estimators trained on TAD-66K and KonIQ-10k correspondingly. 
Based on our analysis, the KonIQ-based classifier aligns more closely with human judgment for high-scoring images. 
Consequently, we apply a stricter threshold for KonIQ compared to the TAD-66K-based model, which shows less consistent performance for top-tier samples.


Following data filtration using non-diffusion-based estimators, we fine-tuned the baseline Stable Diffusion models referenced in our main text. 
We then evaluated these fine-tuned models through side-by-side (SbS) comparisons with their corresponding Alchemist-tuned counterparts. 
The results of this evaluation are presented in Table \ref{tab:filtraion-ablation}.

\begin{table}[htbp]
    \centering
    {
        \begin{tabular}{lccccc}
  \toprule
  \thead[c]{\multirow{2}{*}{Model}} 
  & \multicolumn{4}{c}{\textbf{Side-by-Side Win Rate}} 
   \\
  \cmidrule(lr){2-5} 
  & Rel.$\uparrow$  
  & Aes. $\uparrow$
  & Comp. $\uparrow$ 
  & Fidel. $\uparrow$ \\
  \midrule
  \textbf{SD1.5-Alchemist}
  & & & & & \\
  \quad vs IC9600-TAD66k-KonIQ-sorted
  & \good{0.78} 
  & \neutral{0.54} 
  & \neutral{0.52}
  & \good{0.62} 
  \\

  \cmidrule(lr){1-1} \cmidrule(lr){2-5}
  \textbf{SD2.1-Alchemist}
  & & & & & \\
  \quad vs IC9600-TAD66k-KonIQ-sorted
  & \good{0.80} 
  & \neutral{0.53} 
  &  \good{0.59} 
  &  \good{0.68} 
  \\

  \cmidrule(lr){1-1} \cmidrule(lr){2-5}
  \textbf{SDXL-Alchemist}
  & & & & & \\
  \quad vs IC9600-TAD66k-KonIQ-sorted
  & \good{0.84} 
  & \good{0.63} 
  &  \good{0.58} 
  &  \good{0.68} 
  \\

  \cmidrule(lr){1-1} \cmidrule(lr){2-5}
  \textbf{SD3.5M-Alchemist}
  & & & & & \\
  \quad vs IC9600-TAD66k-KonIQ-sorted
  & \good{0.94} 
  & \neutral{0.46} 
  &  \neutral{0.46} 
  &  \good{0.79} 
  \\

  \cmidrule(lr){1-1} \cmidrule(lr){2-5}
  \textbf{SD3.5L-Alchemist}
  & & & & & \\
  \quad vs IC9600-TAD66k-KonIQ-sorted
  & \good{0.91} 
  & \good{0.62} 
  &  \neutral{0.52} 
  &  \good{0.72} 
  \\

  \bottomrule
\end{tabular}

    }
    \vspace{4pt}
    \caption{Comparison of Alchemist-tuned models against models tuned on the dataset filtrated using TOPIQ-IAA and IC9600, TAD-66k and KonIQ-10k trained classifiers.
    The table reports human win rates (by aspect) w.r.t. Alchemist-tuned models.
    \good{Green} indicates statistically significant improvement ($p < 0.05$), \neutral{gray} no statistically significant change, and \bad{red} a statistically significant decline.}
    \label{tab:filtraion-ablation}
\end{table}

The newly obtained models exhibit two key limitations: (1) reduced image-text coherence and (2) reduced fidelity. 
We attribute these effects to several factors:
\begin{enumerate}
    \item The IC-9600-trained classifier retains excessively complex images in its top selections, whereas our diffusion-based estimator effectively identifies samples with "moderate" complexity - a key characteristic for improving generation quality. Training on overly complex data consistently degrades output fidelity.
    \item Overly strict thresholds on both TAD-66k and KonIQ-10k filters introduce significant content bias in the dataset, ultimately compromising text-to-image alignment during generation.
    \item Visual analysis of the dataset, along with side-by-side (SbS) model comparisons after tuning on this data, shows an important limitation. 
    Existing classifiers are not able to reliably tell apart average-quality images from the aesthetically outstanding samples needed for successful SFT.
\end{enumerate}

This ablation study shows that using TOPIQ-IAA and classifiers trained on TAD-66k, KonIQ-10k and IC9600 does not lead to Alchemist-level integral quality of fine-tuned models.

\subsection{LAION-Aesthetics Size}
\label{appendix:laion_size}
In our primary analysis, we compared the 3,350-sample Alchemist dataset against an equally sized random subset of LAION-Aesthetics v2 \cite{laion-aesthetics-v2} images meeting our minimum size threshold (area $\geq 1024 \times 1024$ px).
In this subsection we validate that the sample size was not the reason for inferior performance of the LAION-based finetuning.

To ablate the influence of the dataset size we select a complete set of 31k samples from LAION-Aesthetics v2 that pass resolution-based selection.
Consistently with our previous fine-tuning experiments, we performed a hyperparameter sweep to train the top-performing models for this dataset. 
After that, we conducted side-by-side (SbS) comparisons against Alchemist fine-tuned versions (Table \ref{tab:laion-size-ablation}).

\begin{table}[htbp]
    \centering
    {
        \begin{tabular}{lccccc}
  \toprule
  \thead[c]{\multirow{2}{*}{Model}} 
  & \multicolumn{4}{c}{\textbf{Side-by-Side Win Rate}} 
   \\
  \cmidrule(lr){2-5} 
  & Rel.$\uparrow$  
  & Aes. $\uparrow$
  & Comp. $\uparrow$ 
  & Fidel. $\uparrow$ \\
  \midrule
  \textbf{SD1.5-Alchemist}
  & & & & & \\
  \quad vs full LAION-tuned
  & \neutral{0.55} 
  & \good{0.59} 
  & \good{0.77}
  & \neutral{0.54} 
  \\

  \cmidrule(lr){1-1} \cmidrule(lr){2-5}
  \textbf{SD2.1-Alchemist}
  & & & & & \\
  \quad vs full LAION-tuned
  & \neutral{0.54} 
  & \good{0.62} 
  & \good{0.76} 
  & \good{0.63} 
  \\

  \cmidrule(lr){1-1} \cmidrule(lr){2-5}
  \textbf{SDXL-Alchemist}
  & & & & & \\
  \quad vs full LAION-tuned
  & \neutral{0.54} 
  & \good{0.66} 
  & \good{0.86} 
  & \good{0.63} 
  \\

  \cmidrule(lr){1-1} \cmidrule(lr){2-5}
  \textbf{SD3.5M-Alchemist}
  & & & & & \\
  \quad vs full LAION-tuned
  & \neutral{0.55} 
  & \good{0.65} 
  & \good{0.82} 
  & \neutral{0.52} 
  \\

  \cmidrule(lr){1-1} \cmidrule(lr){2-5}
  \textbf{SD3.5L-Alchemist}
  & & & & & \\
  \quad vs full LAION-tuned
  & \neutral{0.52} 
  & \good{0.62} 
  & \good{0.72} 
  & \good{0.60} 
  \\

  \bottomrule
\end{tabular}
    }
    \vspace{4pt}
    \caption{Comparison of Alchemist-tuned models and models tuned on the full LAION-Aesthetics v2 dataset. 
    The table reports human win rates (by aspect) w.r.t. Alchemist-tuned models.
    \good{Green} indicates statistically significant improvement ($p < 0.05$), \neutral{gray} no statistically significant change, and \bad{red} a statistically significant decline.}
    \label{tab:laion-size-ablation}
\end{table}

Human evaluation results demonstrate that models trained on the full LAION-Aesthetics v2 dataset continue to underperform those fine-tuned with Alchemist, particularly in measures of aesthetic quality and image complexity.

\section{Experimental Setting}
\label{appendix:experimental_setting}
\subsection{Hyperparameter Sweep and Train Setting}
\label{appendix:sweep}

\begin{table}[htb]
\centering
\begin{tabular}{l|ccc}
\toprule
Model  & Learning Rates & Iterations (thousands)  & EMA $\beta$  \\
\midrule
SD1.5 & [1e-5, 2.5e-5, 8e-5] & [2.5, 5, 7.5, 10, 12.5]  & [0.999, 0.9999] \\ 
SD2.1 & [1e-5, 2.5e-5, 8e-5] & [2.5, 5, 7.5, 10, 12.5] & [0.999, 0.9999] \\
SDXL & [1e-5, 2.5e-5, 8e-5] & [5, 10, 15, 20]  & [0.999, 0.9999]  \\ 
\midrule
SD3.5 M & [5e-6, 2.5e-5, 8e-5] & [20, 40, 60, 80] & [0.9999] \\ 
SD3.5 L &  [1e-6, 5e-6, 2.5e-5] & [20, 40, 60] & [0.9999] \\ 
\bottomrule
\end{tabular}
\vspace{0.1cm}
\caption{Hyperparameter grids during our sweep. The particular choices were made according to the community best practices as well as our computational and human resource constraints.}
\label{tab:sweep_parameters_choice}
\end{table}

\begin{figure}[htbp]
\centering
\begin{subfigure}[b]{0.49\textwidth}
    \includegraphics[width=\textwidth]{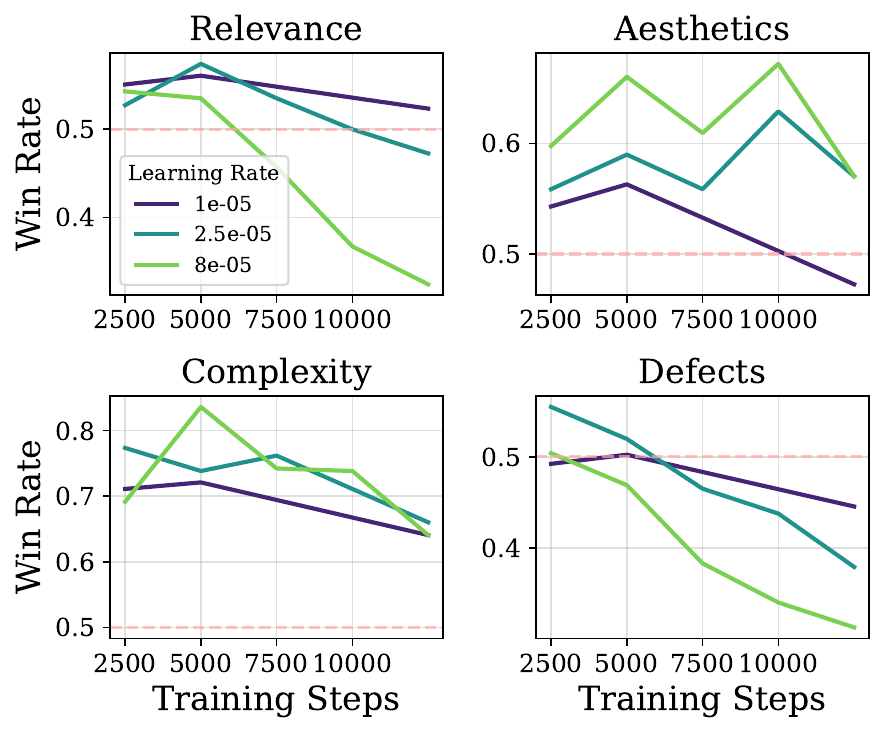}
    \label{fig:sd1.5_dynamics}
    \vspace{-0.5cm}
    \caption{SD1.5.}
    \vspace{0.2cm}
\end{subfigure}
\hfill 
\begin{subfigure}[b]{0.49\textwidth}
    \includegraphics[width=\textwidth]{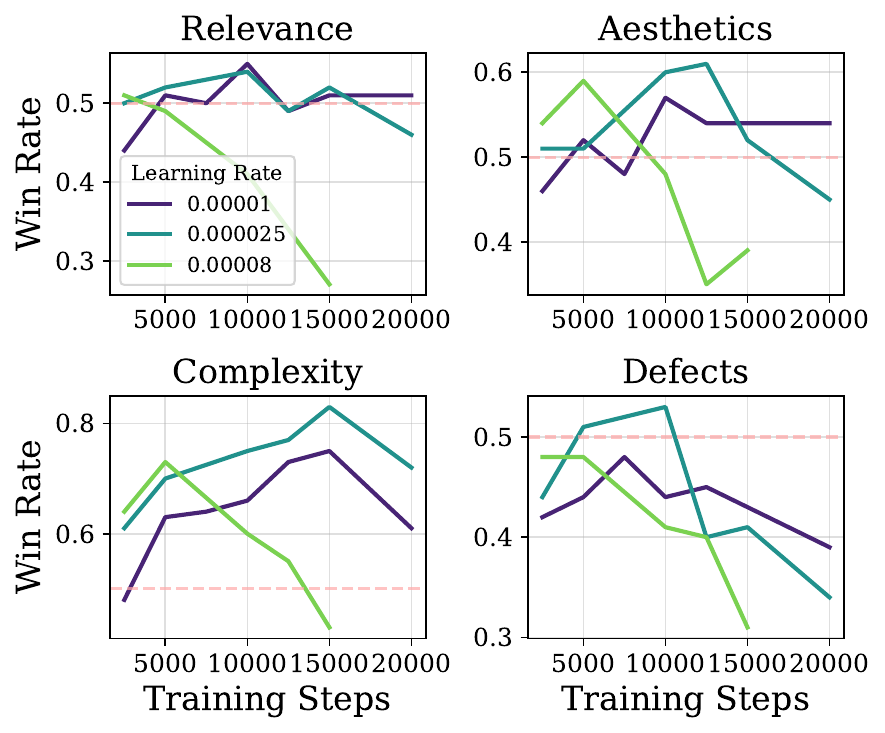}
    \label{fig:sdxl_dynamics}
    \vspace{-0.5cm}
    \caption{SDXL.}
    \vspace{0.2cm}
\end{subfigure}
\begin{subfigure}[b]{0.49\textwidth}
    \includegraphics[width=\textwidth]{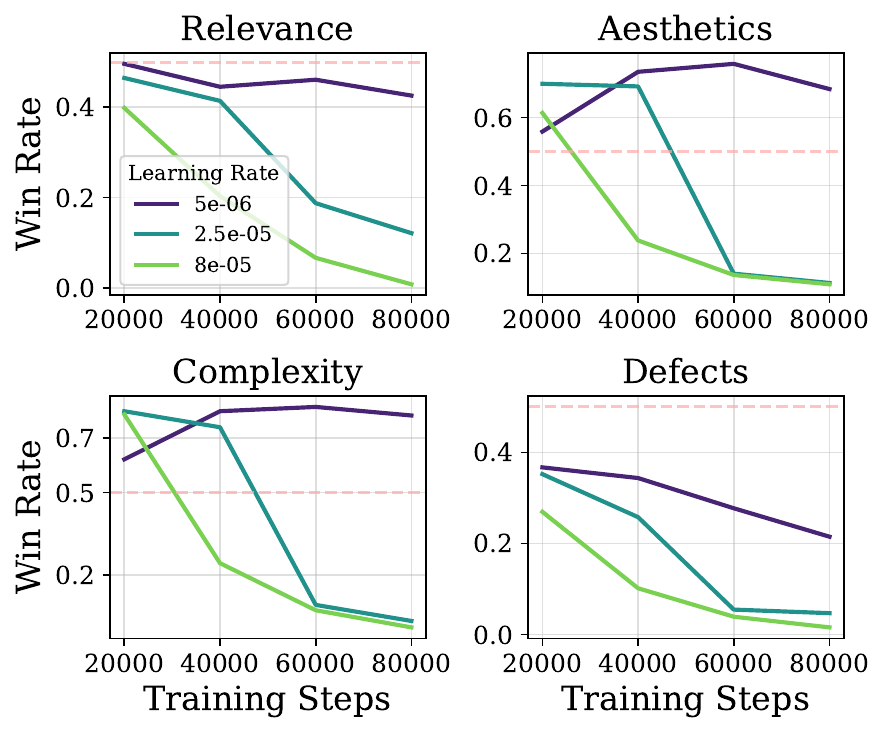}
    \label{fig:sd3.5_m_dynamics}
    \vspace{-0.5cm}
    \caption{SD3.5 Medium.}
    \vspace{0.2cm}
\end{subfigure}
\hfill 
\begin{subfigure}[b]{0.49\textwidth}
    \includegraphics[width=\textwidth]{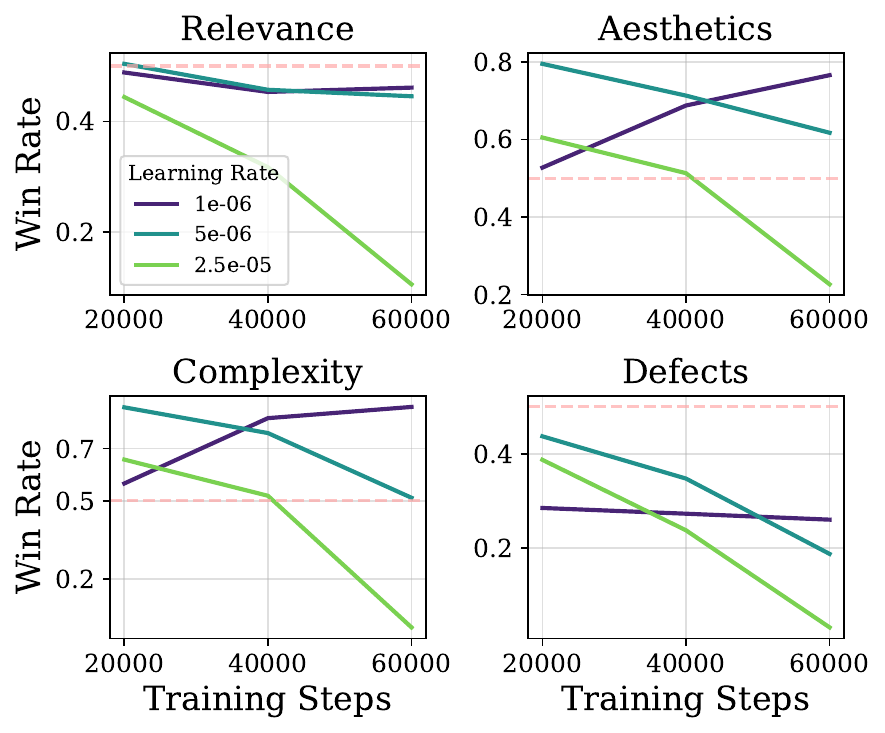}
    \label{fig:sd3.5_l_dynamics}
    \vspace{-0.5cm}
    \caption{SD3.5 Large.}
    \vspace{0.2cm}
\end{subfigure}
\caption{Training dynamics of SD models while tuning on Alchemist. We chose final checkpoints as those maximizing Aesthetics and Complexity while not reaching statistically significant decline in other aspects (if possible at all).}
\label{figure:training_dynamics_for_all}
\vspace{-7px}
\end{figure}

We performed training hyperparameter sweep according to the Table \ref{tab:sweep_parameters_choice} with the resulting training setup presented in the Table \ref{tab:final_training_setup}. Total batch size of 80, AdamW optimizer \cite{loshchilov2017decoupled}, Adam betas $\beta_1=0.9, \beta_2=0.999$ and constant learning rate scheduler were set the same for all the models. We didn't use learning rate warm-up. See Figure \ref{figure:training_dynamics_for_all} for the training dynamics across different learning rates.

\begin{table}[htb]
\centering
\begin{tabular}{l|cccc|cc}
\toprule
Models  & Learning Rate & Iterations & Weight Decay & EMA $\beta$ & GPUs & Mixed Precision \\
\midrule
SD1.5 & 8e-5 & 5k & 1e-2 & 0.999 & 4  &  \texttt{float16}\\ 
SD2.1 & 2.5e-5 & 7.5k & 1e-2 & 0.999 & 4  & \texttt{float16}\\
SDXL & 2.5e-5 & 10k & 1e-2 & 0.999 & 8  & \texttt{float16} \\ 
\midrule
SD3.5 M & 5e-6 & 40k & 1e-4 & 0.9999 & 8  & \texttt{bfloat16}\\ 
SD3.5 L &  5e-6 & 20k & 1e-4 & 0.9999 & 8  & \texttt{bfloat16}\\ 
\bottomrule
\end{tabular}
\vspace{0.1cm}
\caption{Final setup for fine-tuning on the Alchemist.}
\label{tab:final_training_setup}
\end{table}

We used NVIDIA A100 with 80Gb of VRAM, PyTorch 2.6.0 \cite{paszke2019pytorchimperativestylehighperformance}, and CUDA 12.6. 
We varied the number of GPUs from 4 to 8 to ensure the total batch size of 80 on the one hand, and minimize the quantity of GPUs on the other.
Distributed communication is performed via Open MPI \cite{mpi40}. 
We adopt Fully Sharded Data Parallel \cite{zhao2023pytorch} for parameter and optimizer state sharding to reduce memory consumption and allow working with larger batch sizes.

Ultimately, with the final tuning setup it takes 12 GPU-hours to train SD1.5 model, 18 GPU-hours to train SD2.1, 80 GPU-hours to train SDXL, 480 GPU-hours to train SD3.5 Medium and 576 GPU-hours to train SD3.5 Large.

\section{Inference Parameters}
\label{appendix:inference_parameters}

\subsection{Evaluation Setting}
\label{appendix_subsection:eval_setting}

For all models except for the SD3.5 Large we conducted all inference measurements on 1 NVIDIA A100 GPU with 40GB of VRAM, batch size 1, using PyTorch 2.6.0 \cite{paszke2019pytorchimperativestylehighperformance}, and CUDA 12.6. 
For the SD3.5 Large we used the same software, but NVIDIA A100 GPU with 80GB of VRAM.

To generate images we used the parameters either recommended in the corresponding models' HuggingFace repositories or default ones from the Diffusers library. 
These parameters are provided in the Table \ref{tab:default_inference_params}. 
For SDXL we used 80/20\% split of denoising steps between base and refiner models.

\begin{table}[htb]
\centering
\begin{tabular}{l|ccc}
\toprule
Models & Guidance Scale  & Number of Steps & Precision \\
\midrule
SD1.5 & 7.5 & 50 & \texttt{float16} \\ 
SD2.1 & 7.5 & 50 & \texttt{float16} \\ 
SDXL & 5.0 & 50 & \texttt{float16}  \\ 
SD3.5 M & 4.5 & 40 & \texttt{bfloat16} \\
SD3.5 L & 3.5 & 28 & \texttt{bfloat16} \\ 
\bottomrule
\end{tabular}
\vspace{0.1cm}
\caption{Inference parameters used in our work.}
\label{tab:default_inference_params}
\end{table}

\subsection{Inference Parameter Sweep}
\label{appendix_subsection:default_inference}

Although all experiments used default inference parameters, we additionally evaluated model performance across different guidance scales and denoising steps. 
Due to limitations in human evaluation resources, we employed automated assessment using the ImageReward \cite{xu2023imagereward} metric for this analysis.

\begin{figure}[htbp]
\centering
\begin{subfigure}[b]{0.49\textwidth}
    \includegraphics[width=\textwidth]{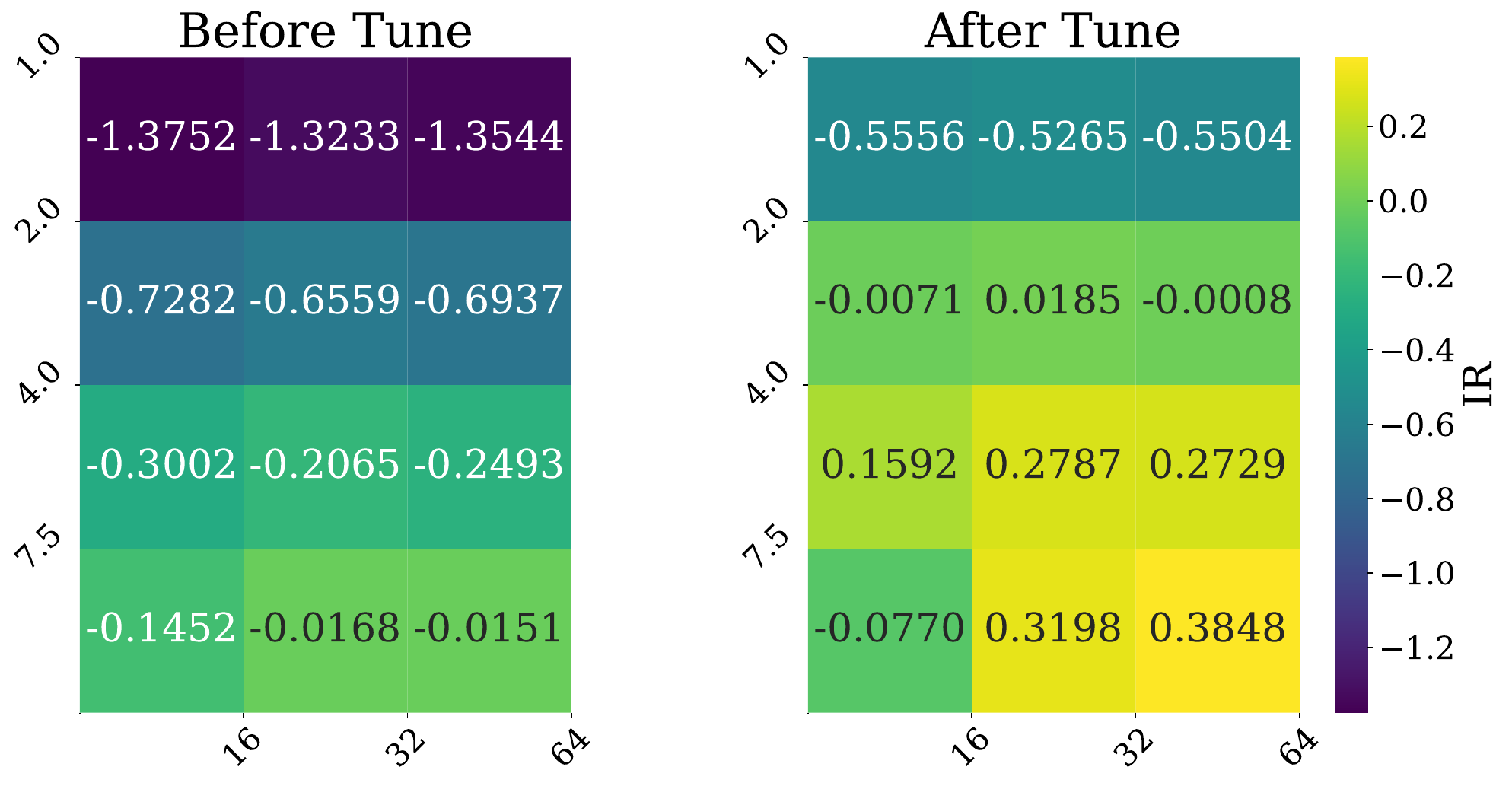}
    \label{fig:sd1.5_IR}
    \vspace{-0.5cm}
    \caption{SD1.5.}
\end{subfigure}
\hfill 
\begin{subfigure}[b]{0.49\textwidth}
    \includegraphics[width=\textwidth]{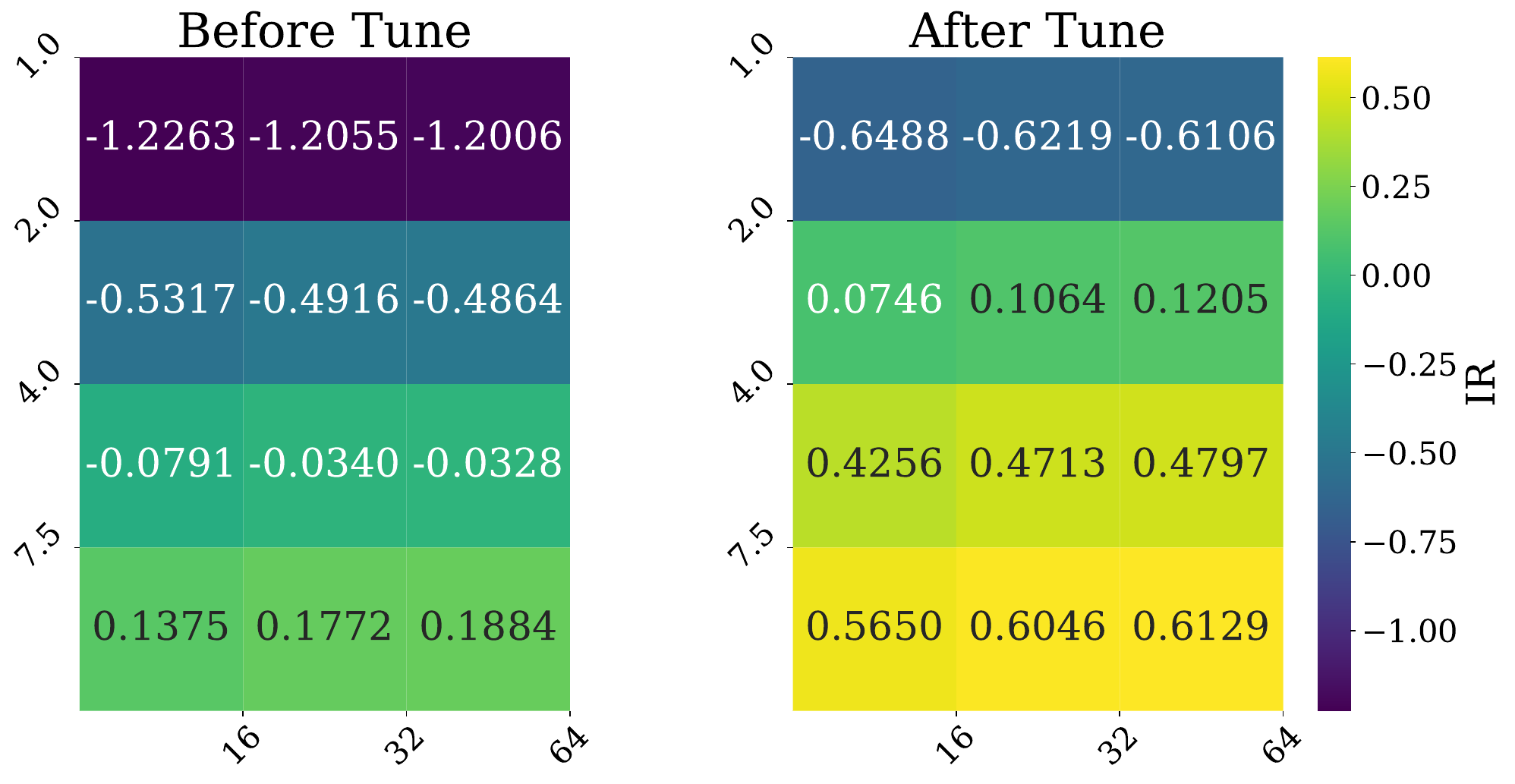}
    \label{fig:sd2.1_IR}
    \vspace{-0.5cm}
    \caption{SD2.1.}
\end{subfigure}
\begin{subfigure}[b]{0.49\textwidth}
    \includegraphics[width=\textwidth]{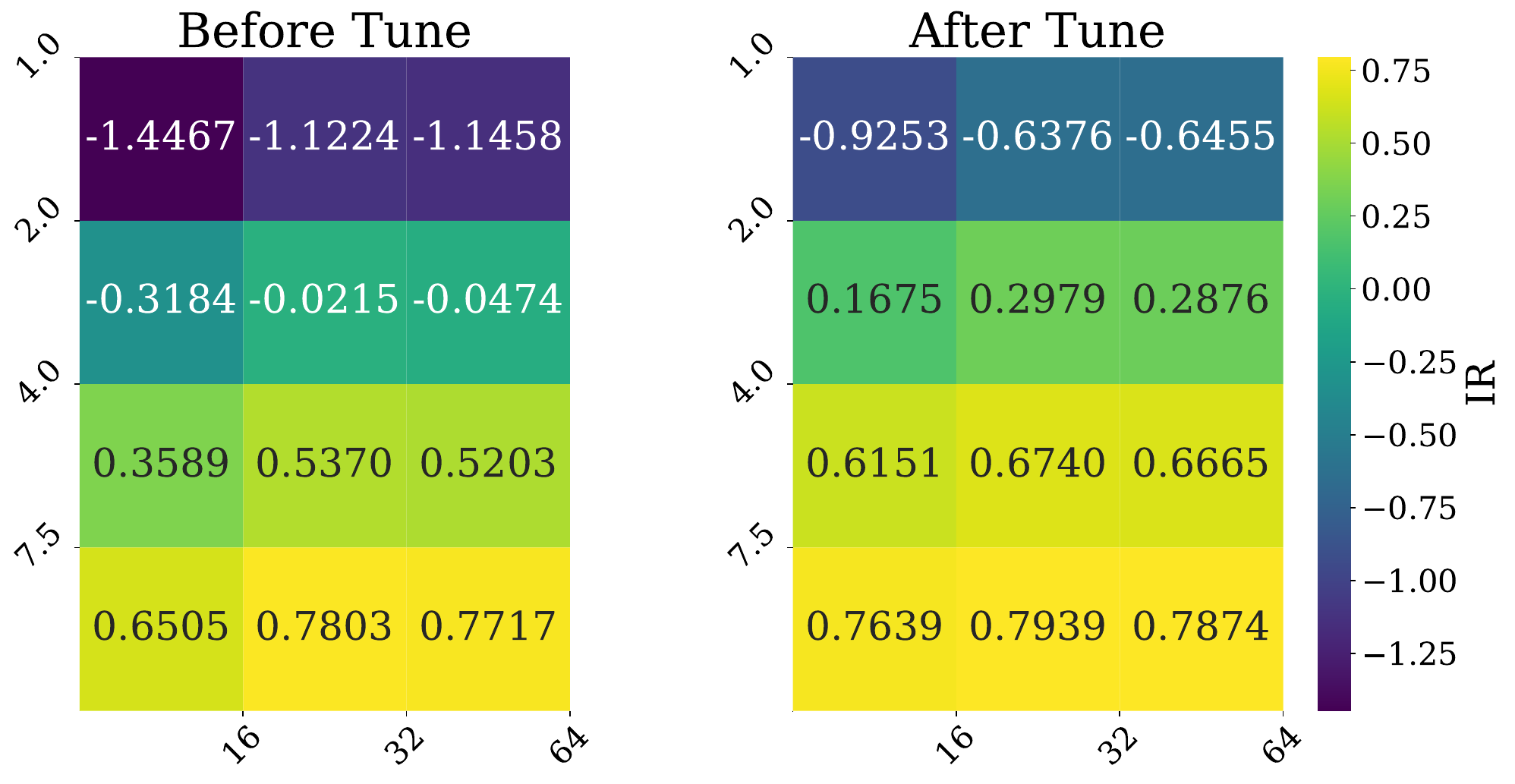}
    \label{fig:sdxl_IR}
    \vspace{-0.5cm}
    \caption{SDXL.}
\end{subfigure}
\hfill 
\begin{subfigure}[b]{0.49\textwidth}
    \includegraphics[width=\textwidth]{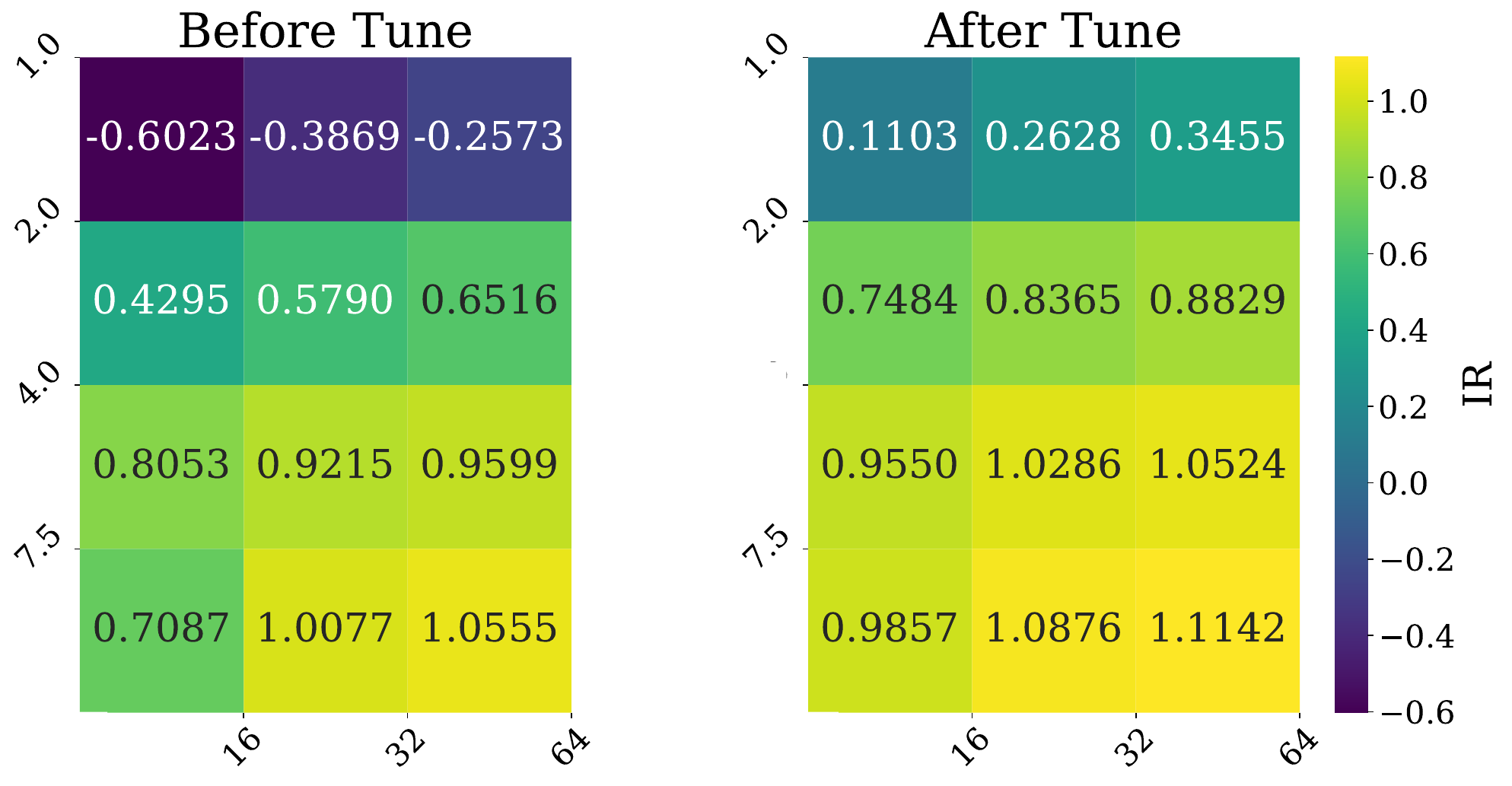}
    \label{fig:sd3.5_IR}
    \vspace{-0.5cm}
    \caption{SD3.5 Medium.}
\end{subfigure}
\caption{ImageReward metric change depending on guidance scale and number of denoising steps \textbf{before} and \textbf{after} tuning on Alchemist.}
\label{figure:inference_heatmaps}
\vspace{-7px}
\end{figure}

Varying guidance scale in $\in [1.0, 2.0, 4.0, 7.5]$ and number of inference steps in  $\in [16, 32, 64]$, we show the dynamics of ImageReward in Figure \ref{figure:inference_heatmaps}. 
Consistent with our primary analysis, all metrics were computed on the MJHQ-30k benchmark dataset \cite{li2024playground}.

Our evaluations demonstrate that Alchemist-based tuning improves overall generation quality, evidenced by increased minimum and maximum ImageReward values. 
However, while the parameter heatmaps reveal ImageReward's preference for higher guidance scales, we caution against using these results as definitive optimization criteria. 
Prior work has established that excessive guidance scales induce overexposure artifacts in generated images \cite{sadat2024eliminatingoversaturationartifactshigh}, suggesting potential limitations in this metric's alignment with human perceptual quality.

\section{Human Evaluation}
\label{appendix:sbs}

We evaluated text-to-image generation quality through controlled side-by-side (SbS) comparisons, where professional assessors selected the superior image for each prompt-image pair. 
Each comparison involved three independent annotations, with final judgments determined by majority vote.

Our evaluation team consists of trained professionals employed under ethical working conditions, including competitive compensation and risk disclosure. 
Assessors have received detailed and fine-grained instructions for each evaluation aspect and passed training and testing before accessing the main tasks. 
We highlight that our organizational equivalent of IRB approved the study.

Annotators evaluated a pair of images generated given the same validation or test prompt based on four criteria:
\begin{itemize}
    \item \textbf{Image-Text Relevance:} Accuracy of the image content relative to the text prompt;
    \item \textbf{Aesthetic Quality:} Overall visual appeal, including composition and style;
    \item \textbf{Image Complexity:} Richness of detail and content within the scene;
    \item \textbf{Fidelity:} Presence and severity of defects, artifacts, distortions, or undesirable elements.
\end{itemize}

\begin{figure}[htpb]
    \centering
    \includegraphics[width=1.0\linewidth]{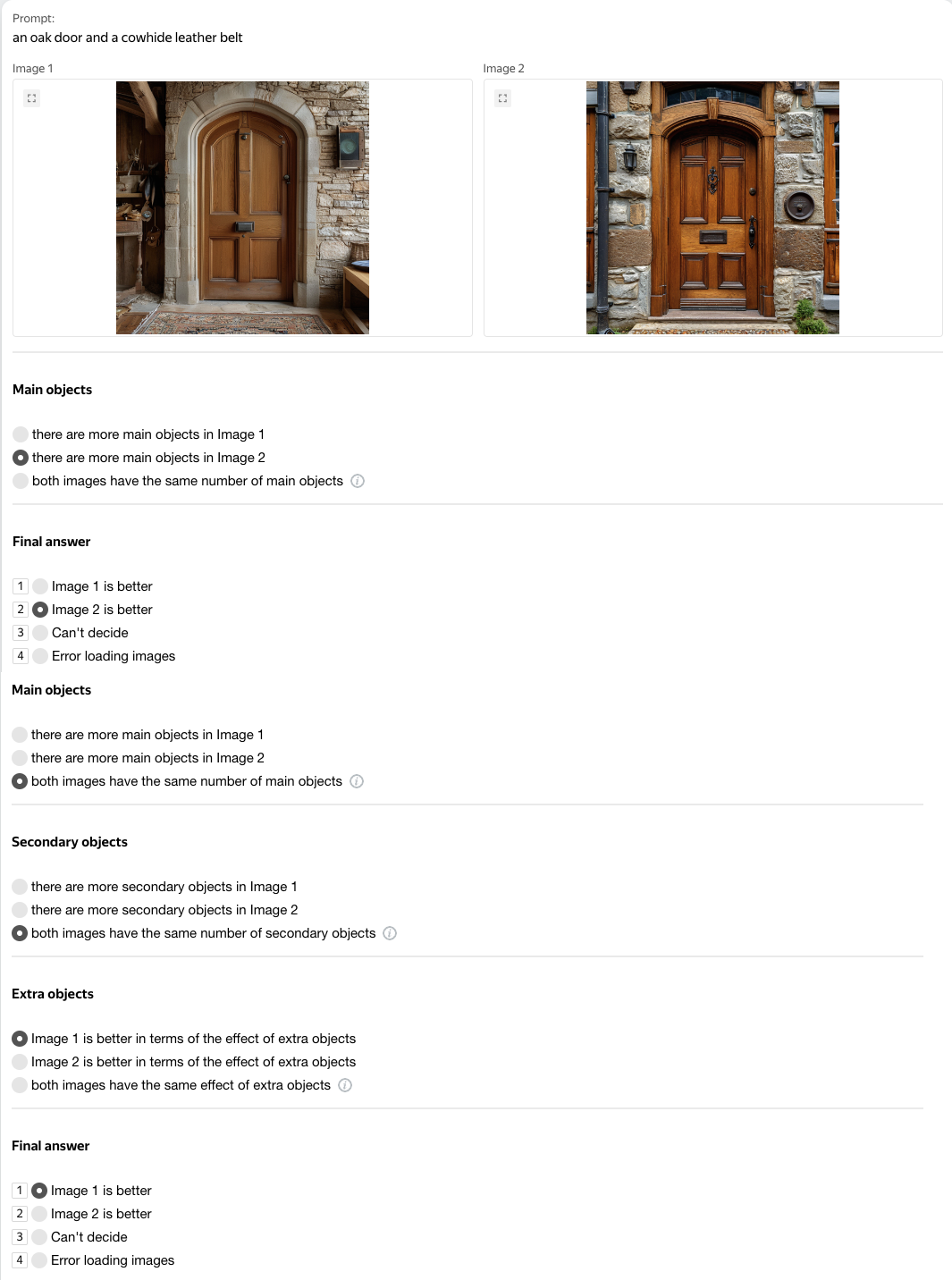}
    \caption{An example of a user interface for the \textbf{Image-Text Relevance} aspect of Human Evaluation with Side-by-Side comparisons.}
    \label{fig:sbs_interface_relevance}
\end{figure}

\begin{figure}[htpb]
    \centering
    \includegraphics[width=1.0\linewidth]{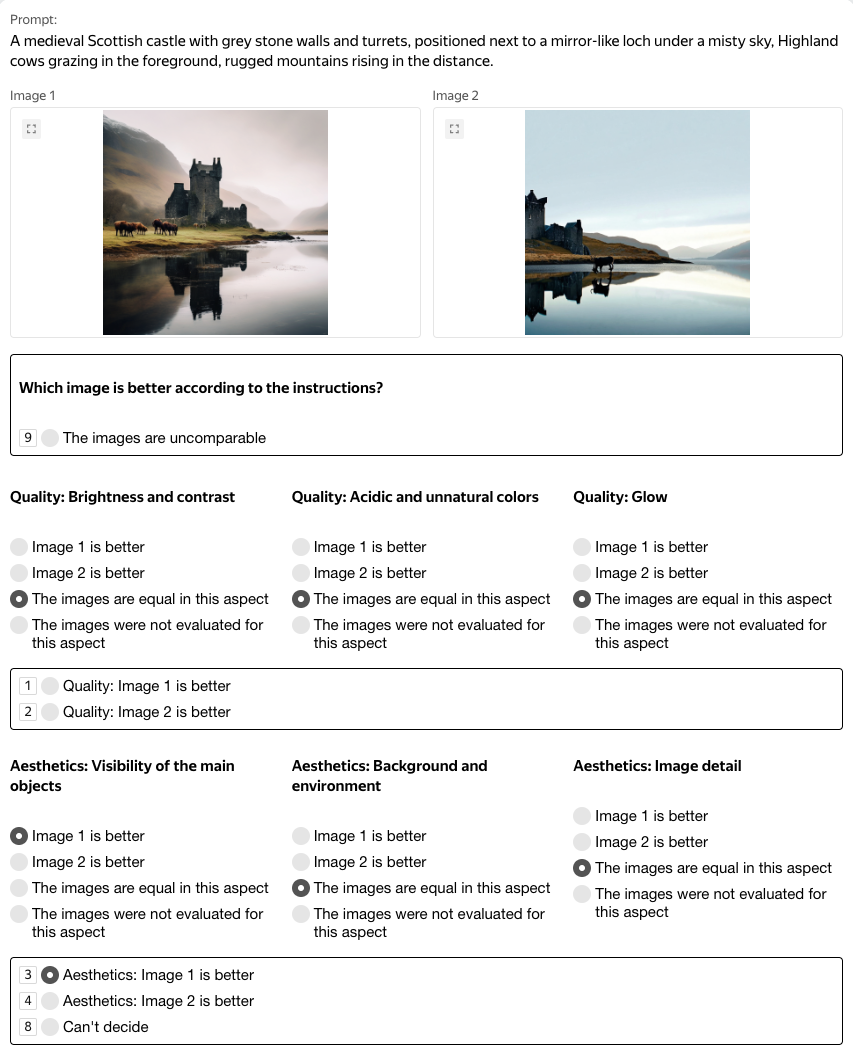}
    \caption{An example of a user interface for the \textbf{Aesthetics} aspect of Human Evaluation with Side-by-Side comparisons.}
    \label{fig:sbs_interface_aesthetic}
\end{figure}

\begin{figure}[htpb]
    \centering
    \includegraphics[width=1.0\linewidth]{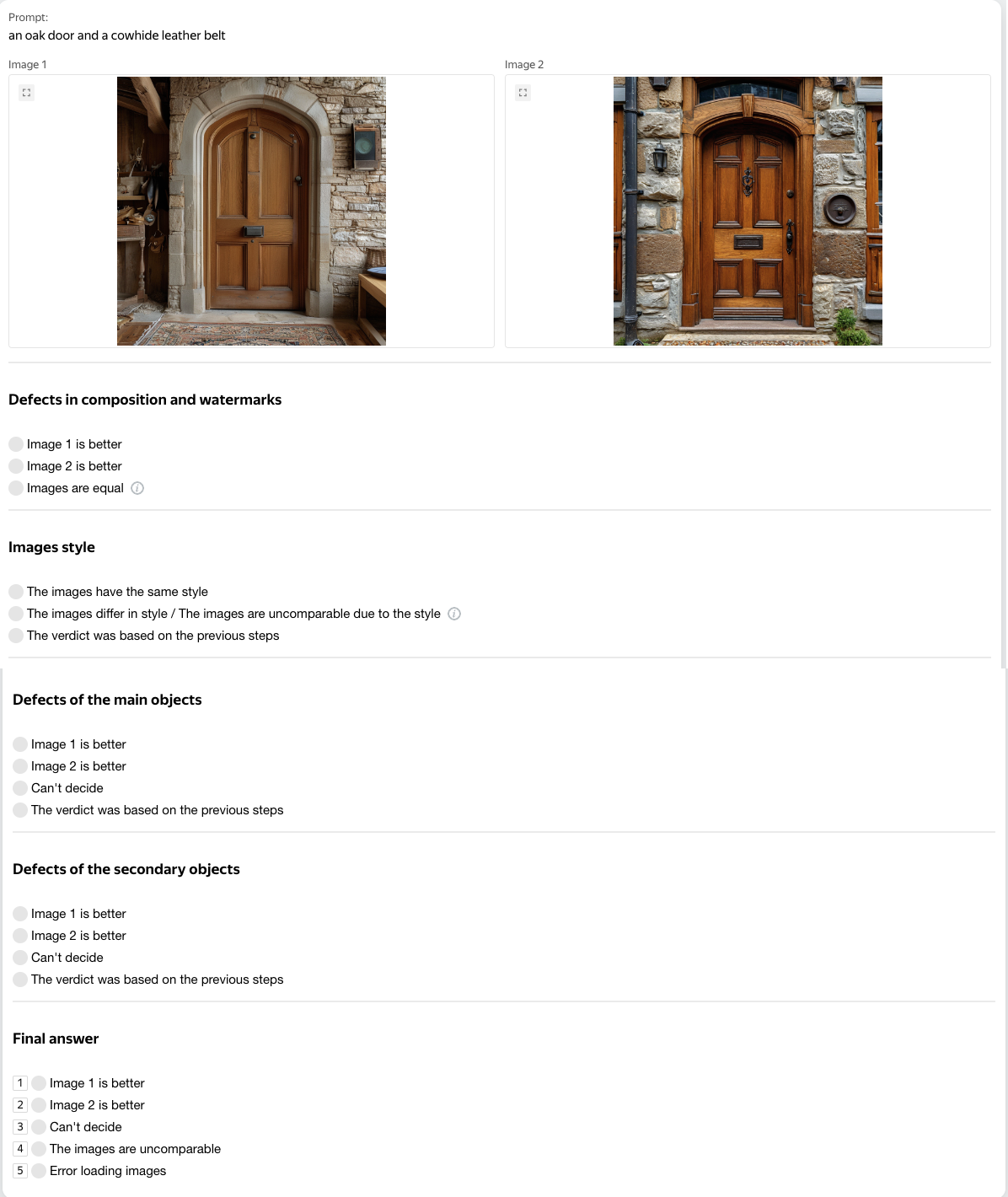}
    \caption{An example of a user interface for the \textbf{Fidelity} aspect of Human Evaluation with Side-by-Side comparisons.}
    \label{fig:sbs_interface_defects}
\end{figure}

\begin{figure}[htpb]
    \centering
    \includegraphics[width=1.0\linewidth]{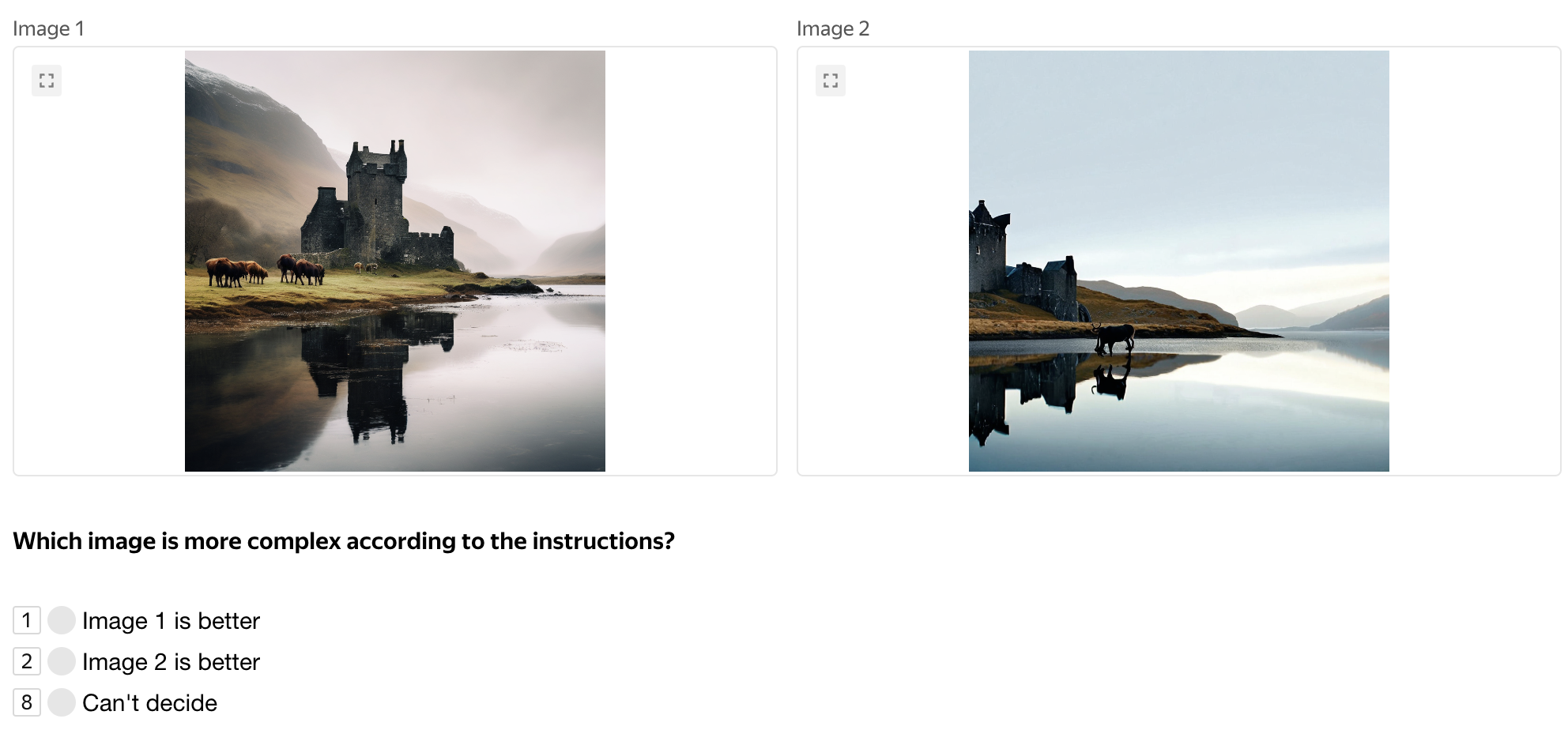}
    \caption{An example of a user interface for the \textbf{Image Complexity} aspect of Human Evaluation with Side-by-Side comparisons.}
    \label{fig:sbs_interface_complexity}
\end{figure}

We provide the platform's interface during each aspect assessment in Figures \ref{fig:sbs_interface_relevance},\ref{fig:sbs_interface_aesthetic},,\ref{fig:sbs_interface_defects},\ref{fig:sbs_interface_complexity}. 
 
From a mathematical point of view, human evaluation is a statistical hypothesis test. 
In particular, we are using a two-sided binomial test and its implementation from \texttt{scipy} \cite{2020SciPy-NMeth} library to test the null hypothesis of whether the two given models are equal in terms of image generation quality in 4 aspects independently. 
More precisely, for each aspect we calculate p-value as following:

\begin{python}
    from scipy.stats import binomtest
    # cnt_wins_baseline   - number of wins for baseline model
    # cnt_wins_experiment - number of wins for experimental model 
    # cnt_equals          - number of equals
    p_value  = binomtest(
        cnt_wins_baseline + cnt_equals / 2, 
        cnt_wins_baseline + cnt_equals + cnt_wins_experiment 
    )
\end{python}

We reject the null hypothesis if  is less than $0.05$, \textit{i.e.}, at the 5\% significance level.

\section{Broader Impact}
\label{appendix:broader_impact}

The release of our open-source SFT dataset and fine-tuned text-to-image diffusion models carries significant societal implications, both positive and challenging. 
By openly sharing these resources, we aim to advance research in generative AI while fostering accessibility and reproducibility. 
The improved aesthetic quality and image complexity offered by our models can empower artists, educators, and small-scale creators, democratizing access to high-quality visual generation tools. 

However, like all generative AI systems, these models present risks that must be carefully managed. 
The potential for misuse-such as generating deceptive imagery or deepfakes-necessitates safeguards, including provenance tracking and responsible deployment practices. 
The environmental impact of training and deploying such models also warrants consideration, encouraging the adoption of efficient fine-tuning techniques and shared computational resources.  

To maximize the benefits of this work while mitigating risks, we emphasize the importance of transparency, collaboration, and oversight. 
Users should disclose AI-generated content where ethically relevant, and developers should engage with diverse stakeholders-including artists and ethicists-to ensure alignment with societal values. 
By proactively addressing these challenges, we hope to contribute to the responsible advancement of generative AI, ensuring that its benefits are widely accessible while minimizing unintended harm.
\newpage

\section{More Visualizations}
\label{appendix:more_vis}

We provide more examples of images generated by models before and after fine-tune on Alchemist. 
The corresponding prompts are listed prior to the grids of images.


\begin{figure}[htbp]
    \centering
    \begin{subfigure}[b]{0.49\textwidth}
        \includegraphics[width=\textwidth]{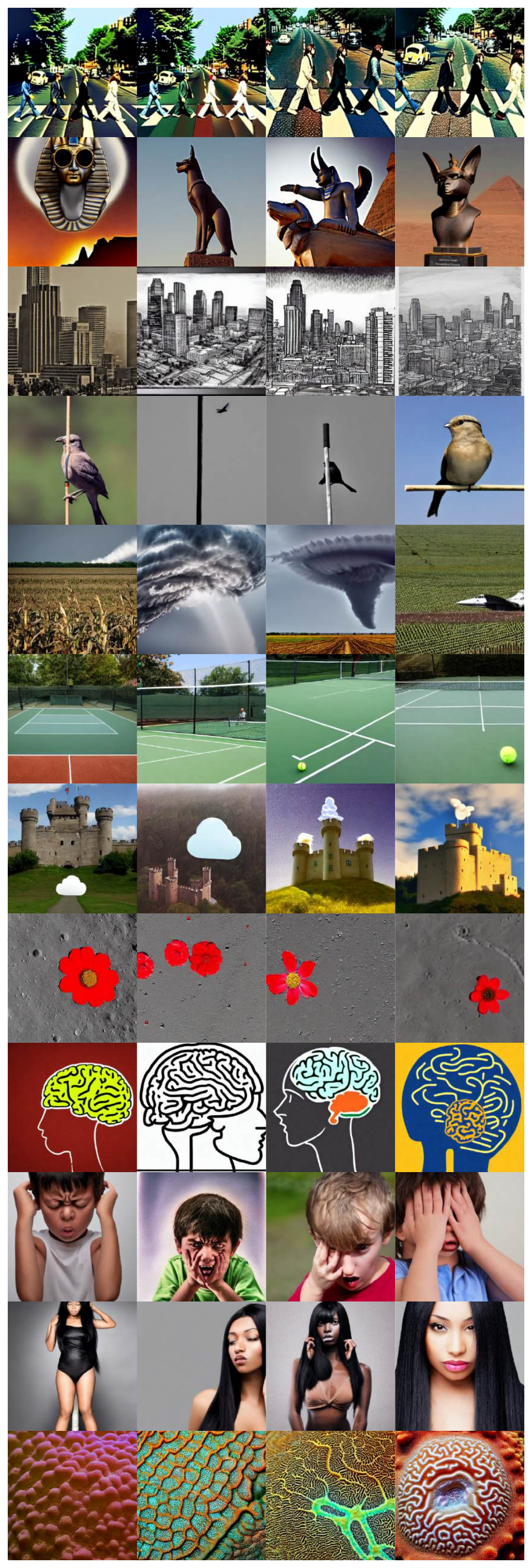}
        \label{fig:sd1_5_extra_before}
    \end{subfigure}
    \hfill 
    \begin{subfigure}[b]{0.49\textwidth}
        \includegraphics[width=\textwidth]{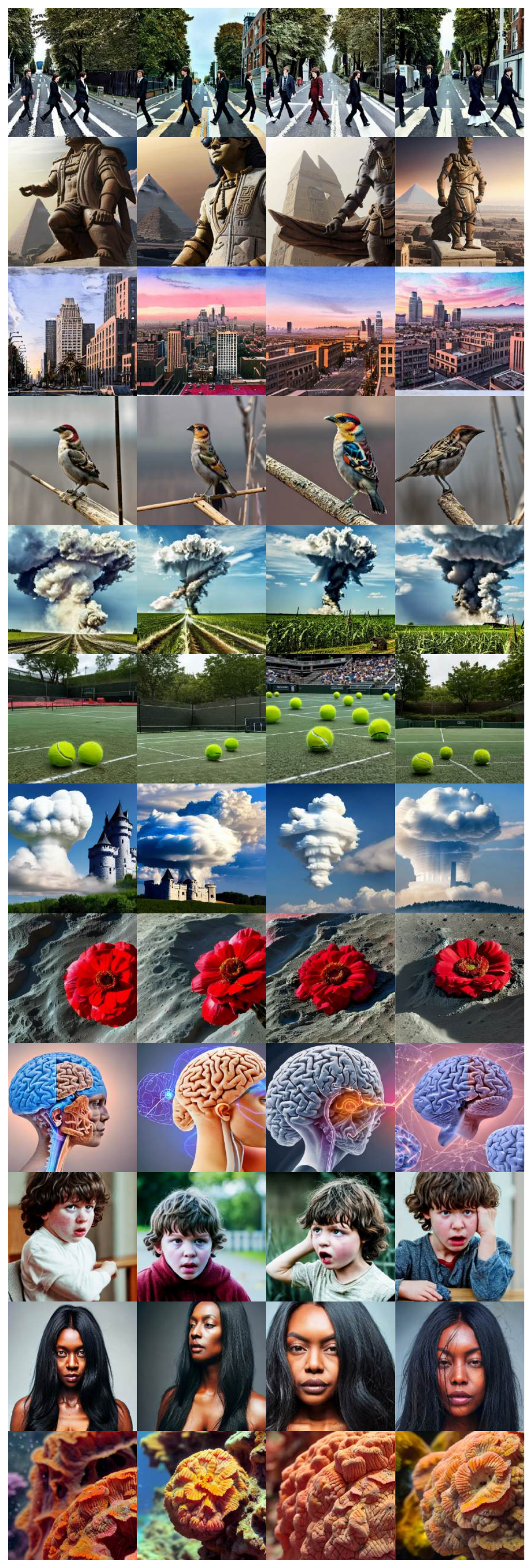}
        \label{fig:sd1_5_extra_after}
    \end{subfigure}
        \caption{More examples of SD1.5 generations \textbf{before} and \textbf{after} tuning on Alchemist. Zoom in for the best view.}
    \label{fig:sd1.5_extra}
\end{figure}

\paragraph{Figure \ref{fig:sd1.5_extra} prompts}
\begin{enumerate}
    \item \textit{"the Beatles crossing Abbey road"}
    \item \textit{"a portrait of a statue of the Egyptian god Anubis wearing aviator goggles, white t-shirt and leather jacket, flying over the city of Mars."}
    \item \textit{"Downtown LA at sunrise. detailed ink wash."}
    \item \textit{"a bird standing on a stick"}
    \item \textit{"a tornado passing over a corn field"}
    \item \textit{"a tennis court with tennis balls scattered all over it"}
    \item \textit{"a cloud in the shape of a castle"}
    \item \textit{"a flower with large red petals growing on the moon's surface"}
    \item \textit{"a diagram of brain function"}
    \item \textit{"a frustrated child"}
    \item \textit{"a woman with long black hair and dark skin"}
    \item \textit{"a macro photograph of brain coral"}
\end{enumerate}
\newpage
\begin{figure}[htbp]
    \centering
    \begin{subfigure}[b]{0.49\textwidth}
        \includegraphics[width=\textwidth]{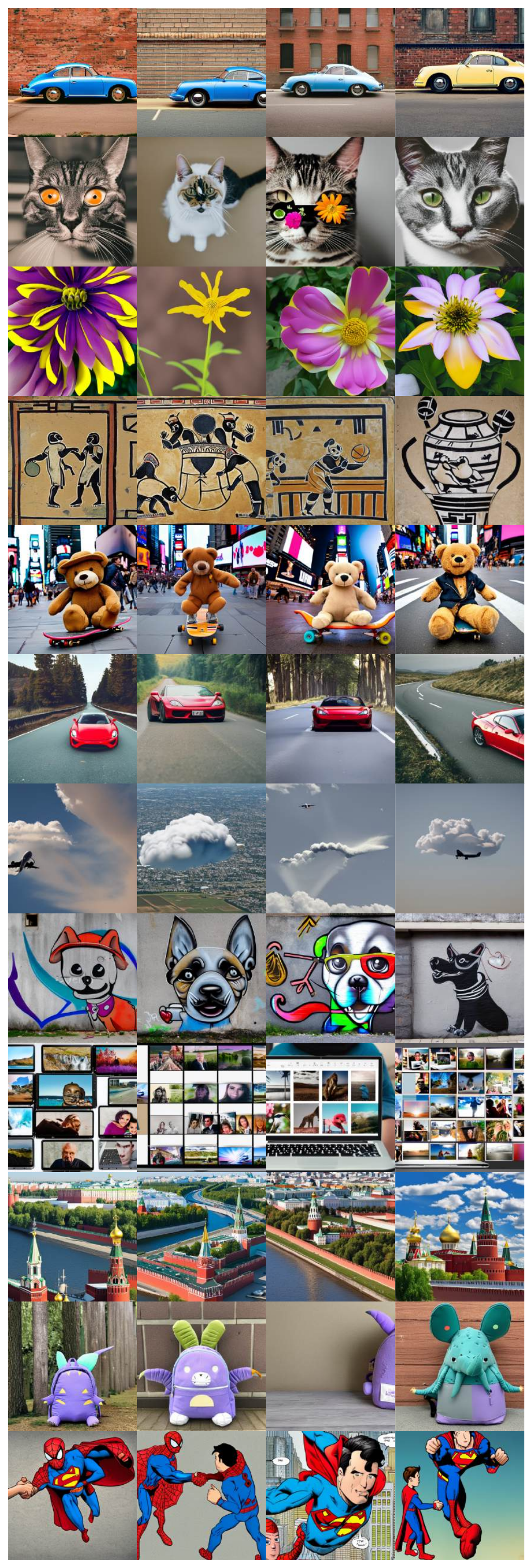}
        \label{fig:sd2.1_extra_before}
    \end{subfigure}
    \hfill 
    \begin{subfigure}[b]{0.49\textwidth}
        \includegraphics[width=\textwidth]{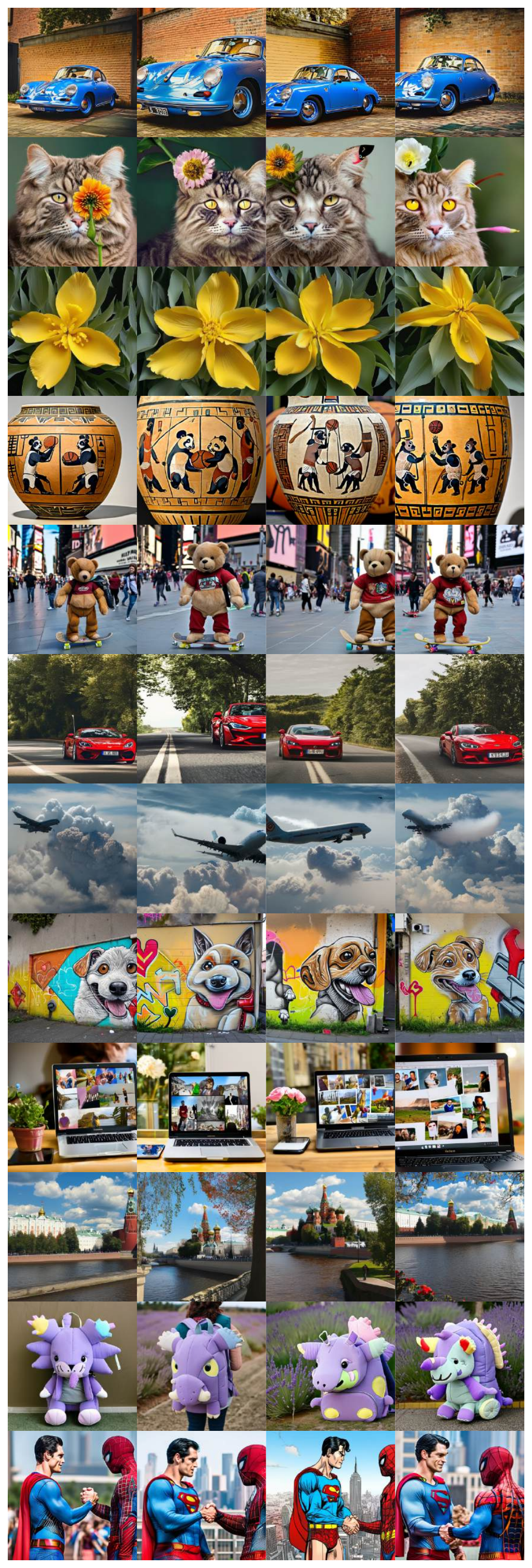}
        \label{fig:sd2.1_extra_after}
    \end{subfigure}
        \caption{More examples of SD2.1 generations \textbf{before} and \textbf{after} tuning on Alchemist. Zoom in for the best view.}
    \label{fig:sd2.1_extra}
\end{figure}

\paragraph{Figure \ref{fig:sd2.1_extra} prompts}
\begin{enumerate}
    \item \textit{"A blue Porsche 356 parked in front of a yellow brick wall"}
    \item \textit{"a flower with a cat's face in the middle"}
    \item \textit{"a flower with large yellow petals"}
    \item \textit{"A photo of an Athenian vase with a painting of pandas playing basketball in the style of Egyptian hieroglyphics."}
    \item \textit{"a teddy bear on a skateboard in times square"}
    \item \textit{"a red sports car on the road"}
    \item \textit{"an airplane flying into a cloud that looks like monster"}
    \item \textit{"graffiti of a funny dog on a street wall"}
    \item \textit{"a laptop screen showing a bunch of photographs"}
    \item \textit{"a view of the Kremlin on a sunny day"}
    \item \textit{"a lavender backpack with a triceratops stuffed animal head on top"}
    \item \textit{"Superman shaking hands with Spiderman"}
\end{enumerate}
\newpage

\begin{figure}[htbp]
    \centering
    \begin{subfigure}[b]{0.49\textwidth}
        \includegraphics[width=\textwidth]{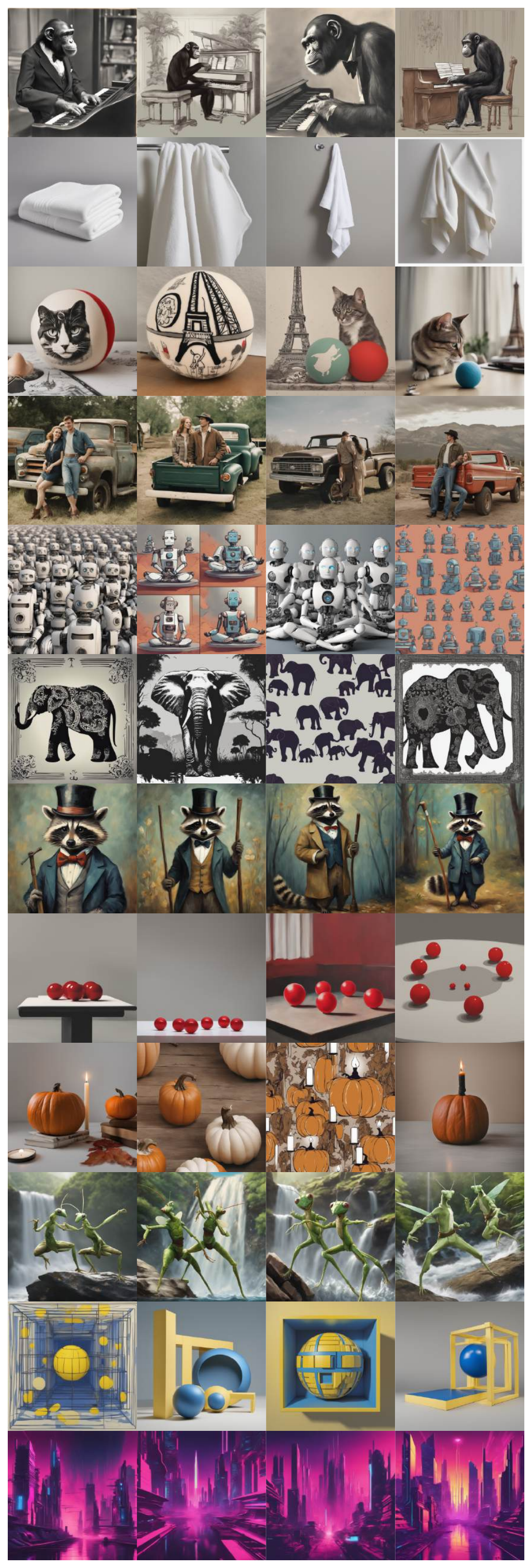}
        \label{fig:sdxl_extra_before}
    \end{subfigure}
    \hfill 
    \begin{subfigure}[b]{0.49\textwidth}
        \includegraphics[width=\textwidth]{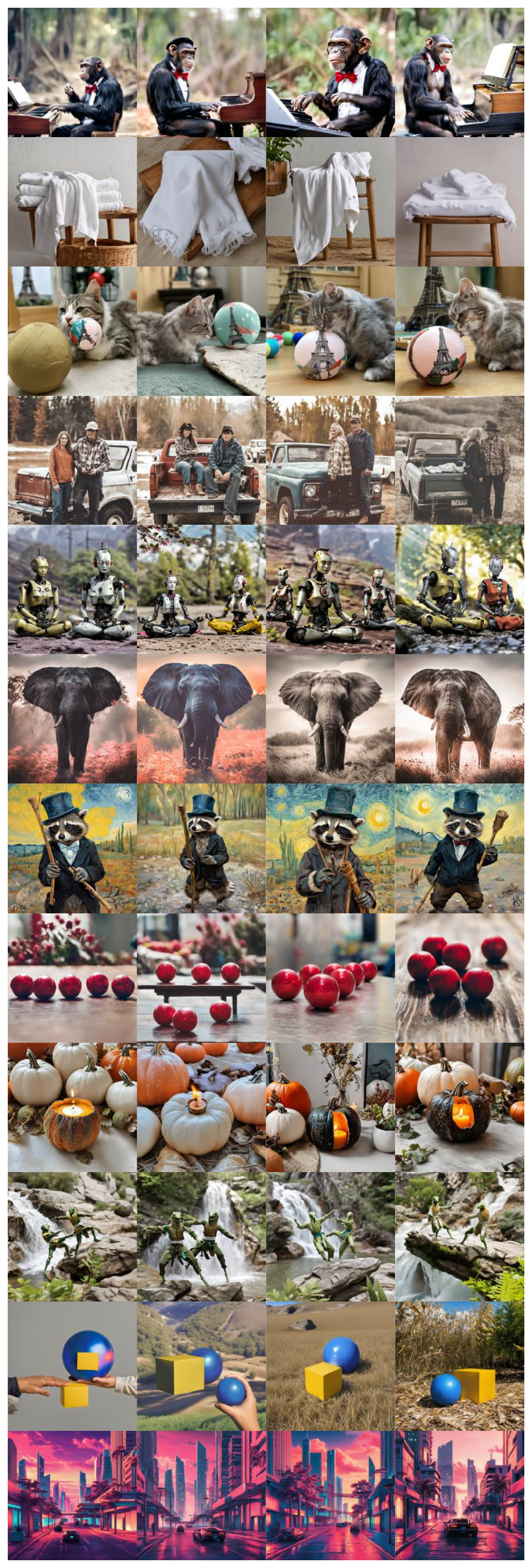}
        \label{fig:sdxl_extra_after}
    \end{subfigure}
        \caption{More examples of SDXL generations \textbf{before} and \textbf{after} tuning on Alchemist. Zoom in for the best view.}
    \label{fig:sdxl_extra}
\end{figure}

\paragraph{Figure \ref{fig:sdxl_extra} prompts}
\begin{enumerate}
    \item \textit{"a chimpanzee wearing a bowtie and playing a piano"}
    \item \textit{"a white towel"}
    \item \textit{"a cat licking a large felt ball with a drawing of the Eiffel Tower on it"}
    \item \textit{"a man and a woman standing in the back up an old pickup truck"}
    \item \textit{"robots meditating"}
    \item \textit{"the silhouette of an elephant"}
    \item \textit{"A raccoon wearing formal clothes, wearing a top hat and holding a cane. The raccoon is holding a garbage bag. Oil painting in the style of Vincent Van Gogh."}
    \item \textit{"five red balls on a table"}
    \item \textit{"a pumpkin with a candle in it"}
    \item \textit{"A close-up of two mantis wearing karate uniforms and fighting, jumping over a waterfall."}
    \item \textit{"a yellow box to the right of a blue sphere"}
    \item \textit{"a futuristic city in synthwave style"}
\end{enumerate}
\newpage

\begin{figure}[htbp]
    \centering
    \begin{subfigure}[b]{0.49\textwidth}
        \includegraphics[width=\textwidth]{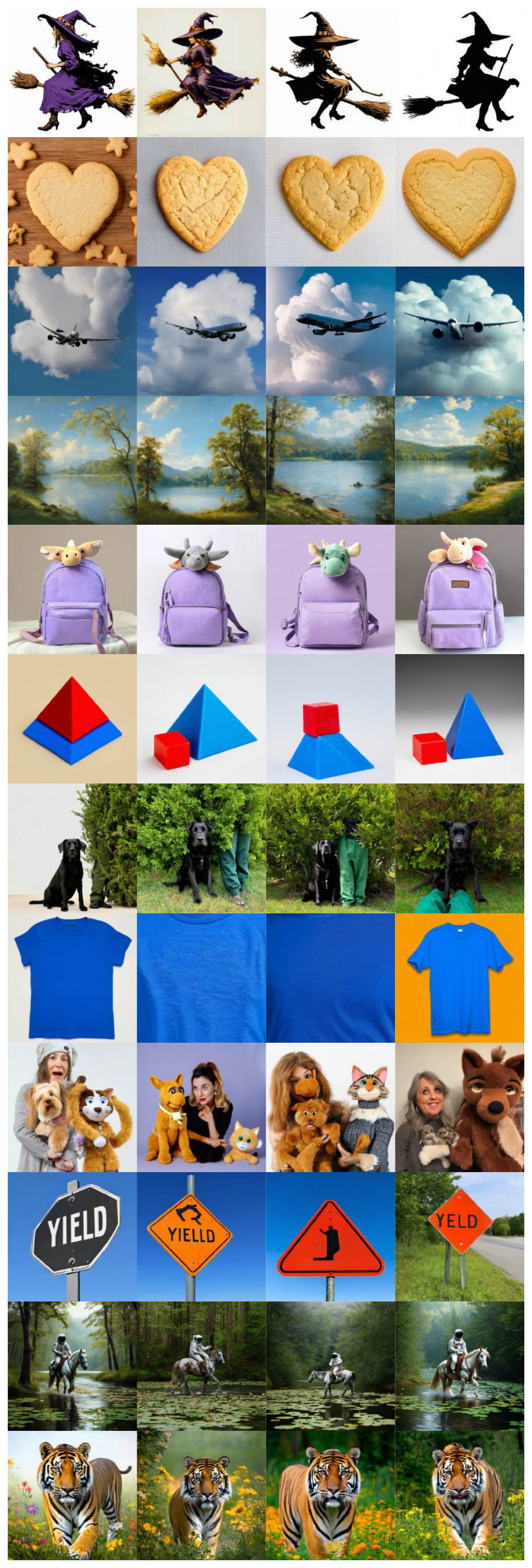}
        \label{fig:sd3.5m_extra_before}
    \end{subfigure}
    \hfill 
    \begin{subfigure}[b]{0.49\textwidth}
        \includegraphics[width=\textwidth]{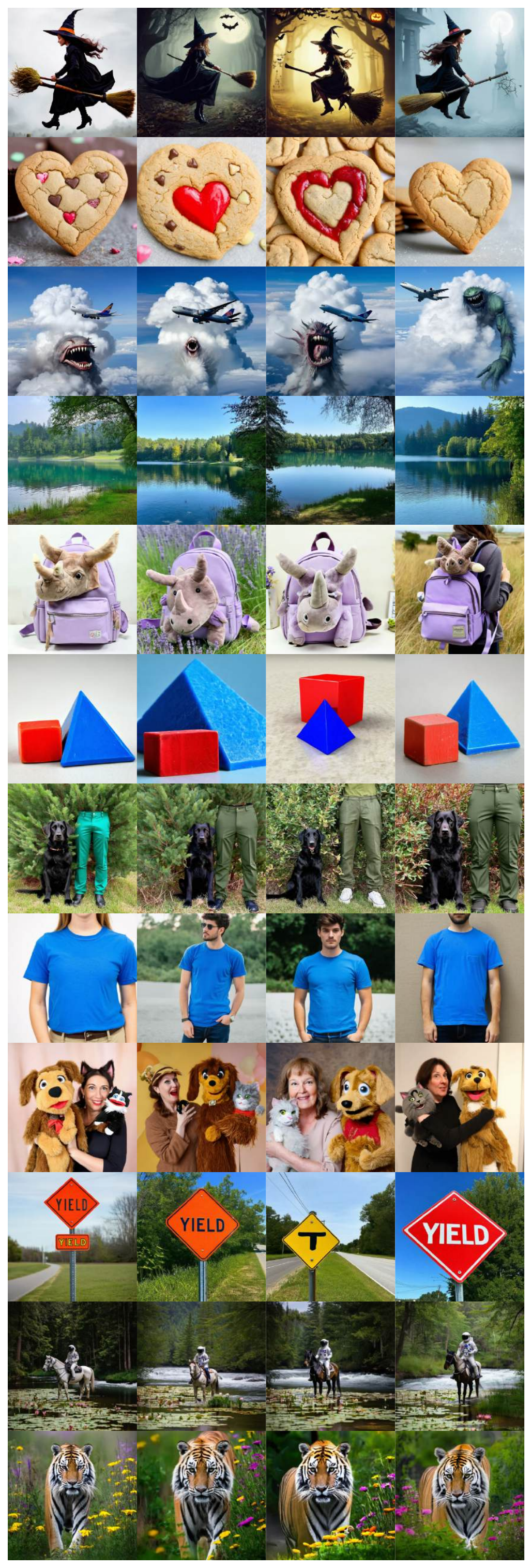}
        \label{fig:sd3.5m_extra_after}
    \end{subfigure}
        \caption{More examples of SD3.5 Medium generations \textbf{before} and \textbf{after} tuning on Alchemist. Zoom in for the best view.}
    \label{fig:sd3.5m_extra}
\end{figure}

\paragraph{Figure \ref{fig:sd3.5m_extra} prompts}
\begin{enumerate}
    \item \textit{"a witch riding a broom"}
    \item \textit{"A heart made of cookie"}
    \item \textit{"an airplane flying into a cloud that looks like monster"}
    \item \textit{"a peaceful lakeside landscape"}
    \item \textit{"a lavender backpack with a triceratops stuffed animal head on top"}
    \item \textit{"a red block to the left of a blue pyramid"}
    \item \textit{"a black dog sitting between a bush and a pair of green pants standing up with nobody inside them"}
    \item \textit{"a blue t-shirt"}
    \item \textit{"a woman with a dog puppet and a cat puppet"}
    \item \textit{"a yield sign"}
    \item \textit{"A photo of an astronaut riding a horse in the forest. There is a river in front of them with water lilies."}
    \item \textit{"a tiger standing by some flowers"}
\end{enumerate}
\newpage

\begin{figure}[htbp]
    \centering
    \begin{subfigure}[b]{0.49\textwidth}
        \includegraphics[width=\textwidth]{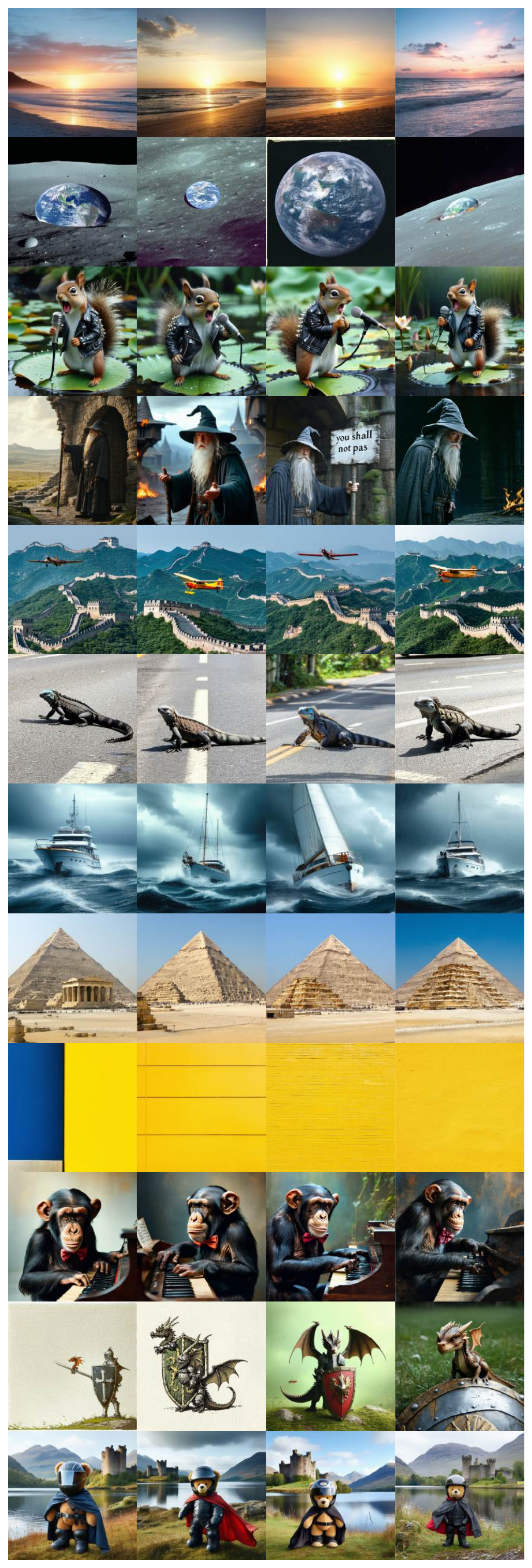}
        \label{fig:sd3.5l_extra_before}
    \end{subfigure}
    \hfill 
    \begin{subfigure}[b]{0.49\textwidth}
        \includegraphics[width=\textwidth]{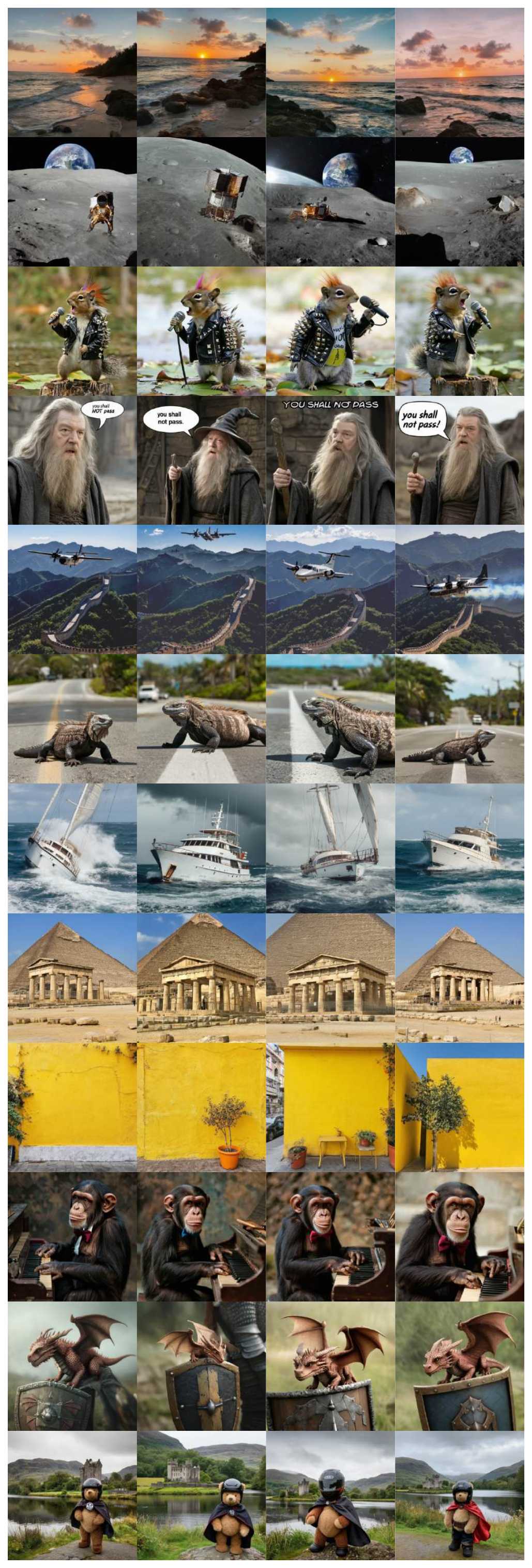}
        \label{fig:sd3.5l_extra_after}
    \end{subfigure}
        \caption{More examples of SD3.5 Large generations \textbf{before} and \textbf{after} tuning on Alchemist. Zoom in for the best view.}
    \label{fig:sd3.5l_extra}
\end{figure}

\paragraph{Figure \ref{fig:sd3.5l_extra} prompts}
\begin{enumerate}
    \item \textit{"The sunset on the beach is wonderful"}
    \item \textit{"a view of the Earth from the moon"}
    \item \textit{"A punk rock squirrel in a studded leather jacket shouting into a microphone while standing on a lily pad"}
    \item \textit{"Gandalf saying you shall not pass"}
    \item \textit{"a prop plane flying low over the Great Wall"}
    \item \textit{"a marine iguana crossing the street"}
    \item \textit{"a large white yacht tossed about in a stormy sea"}
    \item \textit{"the Parthenon in front of the Great Pyramid"}
    \item \textit{"a yellow wall"}
    \item \textit{"a chimpanzee wearing a bowtie and playing a piano"}
    \item \textit{"a tiny dragon landing on a knight's shield"}
    \item \textit{"A teddy bear wearing a motorcycle helmet and cape is standing in front of Loch Awe with Kilchurn Castle behind him"}
\end{enumerate}

\end{document}